\documentclass[11pt]{article}

\usepackage[final]{acl}

\usepackage{amssymb}
\usepackage{amsmath,amsfonts}
\usepackage{tabularx}
\usepackage{booktabs}
\usepackage{graphicx}
\usepackage[percent]{overpic}
\usepackage{algorithm}
\usepackage{algpseudocode}
\usepackage{times}
\usepackage{latexsym}
\usepackage[T1]{fontenc}
\usepackage[utf8]{inputenc}
\usepackage{microtype}
\usepackage{inconsolata}

\newcommand{\notests}{\texttt{no\_tests}}
\newcommand{\known}{\texttt{known}}
\newcommand{\remcts}{ReMCTS}

\title{\remcts{}: Reflection-Enhanced Monte Carlo Tree Search \\for Code Generation}

\author{Huifei Wang\textsuperscript{1}, Xinying Huang\textsuperscript{2}, Yiheng Sun\textsuperscript{1}, Yifan Yuan\textsuperscript{2*} \\
  \textsuperscript{1}College of Computer Science and Software Engineering, Shenzhen University \\
  \textsuperscript{2}School of Artificial Intelligence, Shenzhen University \\
  \texttt{2023040114@email.szu.edu.cn}, \texttt{doris.70huang@gmail.com} \\
  \texttt{2023040450@email.szu.edu.cn}, \texttt{yifanyuan@szu.edu.cn}}

\begin{document}
\raggedbottom

\maketitle
\begingroup
\renewcommand{\thefootnote}{\fnsymbol{footnote}}
\footnotetext[1]{Corresponding author.}
\endgroup

\begin{abstract}

Open-weight large language models (LLMs) can generate function-level programs from natural-language prompts, but plausible candidates still fail on hidden semantics and repeat mistakes across repair attempts. We present \remcts{}, an execution-grounded, memory-augmented, LLM-guided MCTS-style search framework. It organizes program candidates as tree states, retains branch-local debugging context, retrieves failure experience across branches, and distinguishes failed checks from unavailable evidence. On HumanEval and MBPP-Sanitized, visible-test \remcts{} improves over direct generation in 8 of 10 model--dataset pairs under held-out evaluation, whereas proxy-only search is less stable. Controlled tree-search, sampling, repair, and memory ablations characterize the source and limits of these gains. A 30-task HumanEval-X C++ pilot further demonstrates compatibility with compiler-backed execution, but does not constitute a broad multilingual evaluation.
\end{abstract}

\section{Introduction}

Code generation is not merely a one-pass text completion task, but rather a program search problem under partially observable feedback. Even when a model generates syntactically correct code, the program can still fail on boundary conditions, interface constraints, hidden inputs, or complex logic.
These failures make feedback-driven revision a necessary part of the generation process.
This view is close to how human programmers solve coding tasks, in which they write an initial implementation, run it, inspect errors, reflect on the cause, revise the code, and try again. Current LLM code generation often lacks this closed-loop workflow. If the first solution follows a wrong design, later local repair may only patch symptoms while the underlying algorithm remains wrong. Therefore, an effective inference-time system should explore multiple candidate programs, execute them, record failure experience, and reuse that experience across later attempts.
In real-world environments, developers usually verify programs indirectly. Prior work models this practice with compiler feedback \citep{bi2024cocogen}, execution-and-refinement feedback \citep{zheng2025opencodeinterpreter}, unit-test feedback \citep{liu2023rltf}, iterative test-and-repair feedback \citep{tang2026fixaudit}, static-analysis and test loops \citep{ravi2026llmloop}, and interactive feedback for research-code tasks \citep{miao2025recodeh}. These studies suggest that feedback is essential for code generation. Practical open-source code models benefit from feedback when refining candidates over multiple attempts.

Existing approaches improve code generation from several perspectives. Self-Refine \citep{madaan2023self}, Reflexion \citep{shinn2023reflexion}, and Self-Debugging \citep{chen2023selfdebug} repair programs with natural language feedback or execution feedback. LATS \citep{zhou2024language} and MapCoder \citep{islam2024mapcoder} combine language-model reasoning with search, planning, or multi-agent collaboration. CodeRL \citep{le2022coderl}, RLTF \citep{liu2023rltf}, and CODERL+ \citep{jiang2026coderlplus} use unit tests or execution semantics to build reward signals. These studies show that feedback, search, planning, and reflection can improve code generation.
However, existing approaches still leave three gaps.
1) Single-trajectory self-repair can remain trapped in an early wrong algorithmic direction. This occurs because later revisions are conditioned on the same trajectory and often patch local symptoms rather than reconsider alternative algorithms.
2) Tree-search methods often rely on numerical rewards and underuse cross-branch natural-language error experience. Scalar rewards can rank candidates, but they do not preserve the textual diagnoses needed to help sibling branches avoid similar failures.
3) No-test tasks require distinguishing missing evidence from genuine success. Without explicit tests, semantic-only acceptance can be over-optimistic, allowing plausible but weakly checked programs to terminate the search too early.

To address these limitations, we propose \remcts{}, an execution-grounded, memory-augmented, LLM-guided MCTS-style framework for code generation. \remcts{} maintains multiple candidate programs in a search tree instead of repairing one candidate along a single trajectory. The multi-branch design helps the system recover from an early wrong algorithmic direction. Each tree node stores the candidate program, visit statistics, reward information, execution metadata, and path reflection memory. The node states let the search use both program behavior and prior repair experience. A memory-aware selection strategy combines numerical rewards, exploration terms, and semantic memory to allocate the search budget. Sandbox execution and multi-source evidence provide dense rewards that distinguish runnable, partially correct, and risky programs. Failure diagnoses are stored in both path memory and a global memory repository, so later branches can avoid rediscovering the same failure modes. The method inherits the tree policy, UCB selection, and backpropagation structure of MCTS, but uses LLM-generated expansions and no classical random rollout; we use ``MCTS-style'' in this precise sense.

Our contributions are summarized as follows:
\begin{itemize}
    \item We formalize execution-grounded program search in which program candidates are tree states, and structured failure diagnoses become reusable path-level and cross-branch search artifacts. This interface reduces dependence on a single repair trajectory.
    \item We propose an evidence-driven reward and acceptance mechanism. The mechanism integrates execution, proxy, semantic, and risk evidence. The design supports both explicit-test and no-explicit-test settings while distinguishing missing evidence from negative evidence.
    \item We evaluate \remcts{} through controlled comparisons on multiple non-GPT models, separating search feedback from hidden evaluation. Paired tests, budget-up baselines, component ablations, and a limited C++ pilot characterize when the method helps and where it does not.
\end{itemize}

The source code and reproducibility materials are publicly available at
\url{https://github.com/AkieSUKI/ReMCTS}.

\section{Related Work}


\subsection{One-Shot Code Generation and Candidate Selection}

Function-level code generation is often studied as prompt-to-code generation and evaluated with benchmarks such as HumanEval \citep{chen2021evaluating} and MBPP \citep{austin2021program}. Many methods improve this setting by generating multiple candidates and then selecting among them. CodeT \citep{chen2023codet} generates tests and uses test agreement to select code candidates. LEVER \citep{ni2023lever} learns to verify language-to-code outputs with execution results. Coder-Reviewer \citep{zhang2023coderreviewer} reranks generated code with a reviewer model. These methods improve candidate selection, but they mainly operate after candidates have been generated. They do not maintain a shared search state where execution feedback and reflection from failed candidates guide later branches.

\subsection{Feedback-Based Code Repair}

Feedback-based methods repair programs after errors are observed. Self-Refine \citep{madaan2023self} iteratively improves model outputs with self-feedback. Reflexion \citep{shinn2023reflexion} stores verbal feedback from failures, and Self-Debugging \citep{chen2023selfdebug} uses execution feedback and explanation to repair programs. OpenCodeInterpreter \citep{zheng2025opencodeinterpreter} builds execution-and-refinement data for code models. Other work turns tests or execution semantics into reward signals. RLTF \citep{liu2023rltf} extracts rewards from unit-test feedback, and CODERL+ \citep{jiang2026coderlplus} aligns code generation with execution semantics. These studies show that feedback is useful for code generation. However, many repair methods still follow a single trajectory, or use feedback mainly for training, filtering, or reranking. \citet{olausson2024selfrepair} further show that self-repair depends strongly on feedback quality and on the model's ability to use that feedback.

\subsection{Search-Based and Tree-Based Code Generation}

Search and planning methods explore more than one generation path. LATS \citep{zhou2024language} combines language agents with Monte Carlo tree search. MapCoder \citep{islam2024mapcoder} and AgentCoder \citep{huang2023agentcoder} use multiple agents for retrieval, planning, coding, testing, or debugging. PlanSearch \citep{wang2024plansearch} searches over natural-language plans before code generation. S* \citep{li2025sstar} studies test-time scaling for code generation. RethinkMCTS \citep{li2025rethinkmcts} applies Monte Carlo tree search to refine erroneous thoughts in code generation. These methods are closest to \remcts{} because they use search, planning, or test-time compute. \remcts{} instead uses generated programs as the primary search states; each node carries attempted-execution evidence, reward traces, and path reflections, including when a candidate does not compile or run. Diagnoses from one branch can be retrieved by later branches through a task-local global memory repository. Candidate evaluation is also availability-aware: missing tests are marked as unavailable rather than negative evidence. The distinguishing contribution is therefore the integration of program-state tree search, within-task reusable error memory, and availability-aware evidence, rather than tree search alone.

\section{\remcts{} Methodology}
\label{sec:method}

\subsection{Overview}

Given a natural-language task description $P$ and an optional visible test set $\mathcal{T}_{vis}$, code generation's goal is to use a base code model $f_\theta$ to produce a candidate program $S^\star$ that passes the final hidden tests $\mathcal{T}_{hid}$. During search, the system can only access $P$ and the feedback allowed by the setting. We consider two search-feedback settings. In the \known{} setting, the system can use $\mathcal{T}_{vis}$ as execution feedback. In the \notests{} setting, no explicit tests are available. The system then relies on proxy evidence, such as contracts, static checks, automated probes, semantic reviews, and risk scans.

\remcts{} formulates the problem as test-time tree search. The root node corresponds to the original task. Edges correspond to generation or repair actions from the model. Child nodes correspond to new candidate programs. Unlike single-trajectory repair, \remcts{} keeps multiple candidate directions. It reallocates the search budget through reward backpropagation and reflection memory. Each search iteration follows a five-stage loop: selection, expansion, execution, reflection, and backpropagation.
In the \textbf{selection} stage, the system starts from the root and selects a node for expansion with a memory-augmented Upper Confidence Bound score ($UCB_M$). Here $UCB_M$ denotes a memory-aware version of the standard Upper Confidence Bound (UCB) rule. It balances exploitation and exploration while adding semantic memory. In the \textbf{expansion} stage, the code model generates $b$ child candidates. The prompt includes the task, current program, execution feedback, path reflections, and global experience. In the \textbf{execution} stage, the sandbox runs each candidate and collects test, contract, static, probe, semantic, and risk evidence. In the \textbf{reflection} stage, the system generates a structured reflection when a candidate is not accepted. The reflection records the error type, root cause, and repair hint. In the \textbf{backpropagation} stage, the system propagates rewards along the search path and writes generated reflections into the global memory repository.

\begin{figure*}[t]
    \centering
    \begin{overpic}[width=\linewidth]{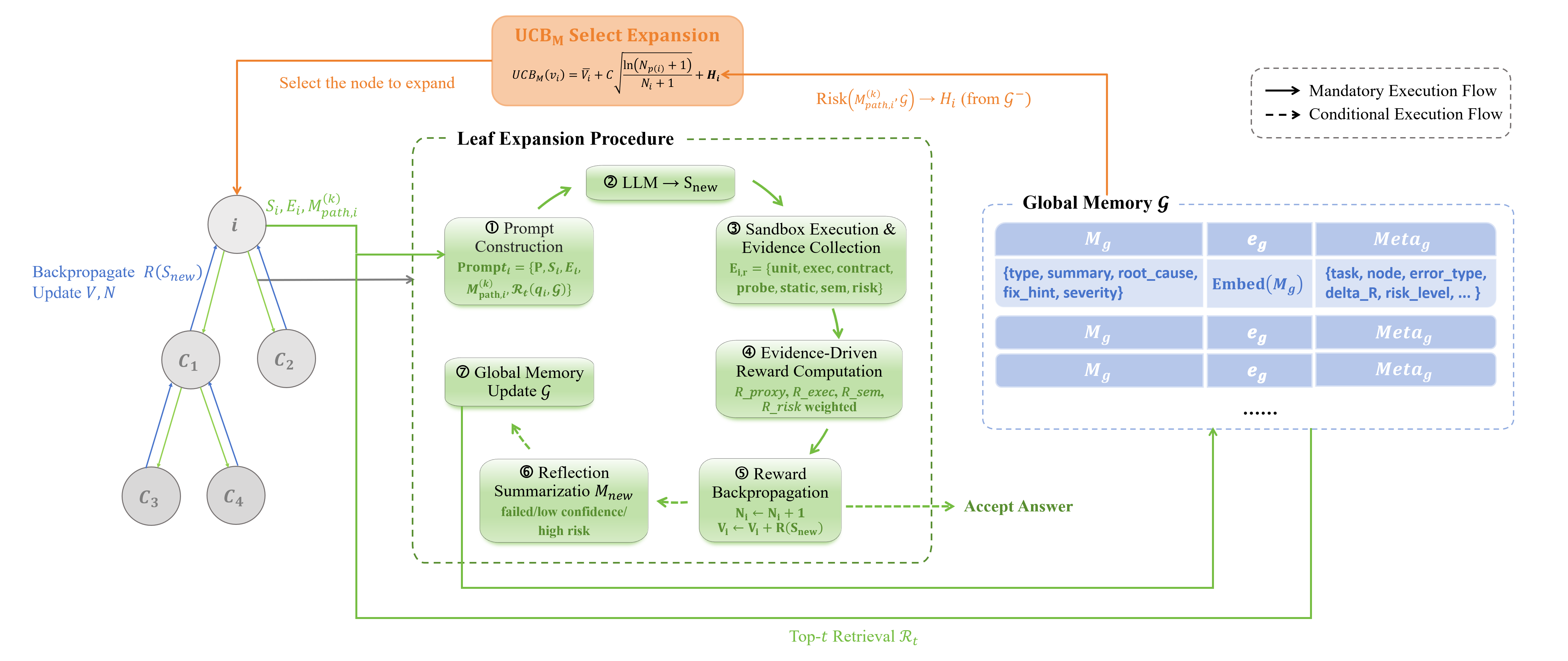}
        \put(37.55,25.70){\makebox(0,0)[lb]{\fontsize{5.4}{5.4}\selectfont n}}
    \end{overpic}
    \caption{Overview of \remcts{}. The search proceeds through selection, expansion, execution, reflection, and backpropagation. Execution evidence determines rewards and acceptance, while path and global reflection memories guide later selection and expansion.}
    \label{fig:overview}
\end{figure*}

Figure~\ref{fig:overview} illustrates the leaf-expansion procedure. Selection, expansion, execution, reflection, and backpropagation form one closed loop. Execution feedback is converted into both numerical rewards and retrievable semantic experience. Reflection affects later node selection and expansion prompts; acceptance is determined by the evidence gate defined below.

\subsection{Memory-Guided Monte Carlo Tree Design}

Each node represents a candidate program. Each node also stores statistical value and semantic history. A node is defined as:
\begin{equation}
    Node_i = [S_i,V_i,N_i,M_{path,i}^{(k)},\Phi_i],
\end{equation}
where $S_i$ denotes the current candidate program. $V_i$ and $N_i$ denote cumulative value and visit count. $\Phi_i$ denotes metadata, such as execution feedback, reward decomposition, risk information, and semantic reviews. The symbol $v_i$ denotes the tree node whose state vector is $Node_i$. Let the path from the root to $v_i$ be $\pi_i=(v_0,\ldots,v_i)$. Let $M_j$ be the structured reflection produced at node $v_j$, when such a reflection exists. The $k$-hop path reflection memory is:
\begin{equation}
M_{path,i}^{(k)}
=\operatorname{Tail}_k\!\left(
(M_j)_{v_j\in\pi_i,\ M_j\neq\varnothing}
\right),
\end{equation}
where $\operatorname{Tail}_k(\cdot)$ denotes the operator that keeps the most recent $k$ reflections. Here $k$ denotes a fixed memory-window hyperparameter. $\varnothing$ denotes the case where no valid reflection is produced at a node. Each valid $M_j$ denotes a structured record with error type, failure symptom, root cause, repair hint, and severity. Thus, a node records two kinds of information: the generated code and recent errors on the path with the corresponding repair hints.

Standard MCTS uses UCB to balance high-value nodes and under-visited nodes. Error patterns in code generation are often semantically repetitive. Examples include entry-point mismatches, missing boundary cases, excessive complexity, and task misunderstanding. To exploit this structure, \remcts{} augments UCB with a \textbf{memory heuristic}:
\begin{equation}
UCB_M(v_i)=\bar{V}_i+
C\sqrt{\frac{\ln(N_{p(i)}+1)}{N_i+1}}+
H_i ,
\end{equation}
where $\bar{V}_i=V_i/(N_i+\epsilon)$ denotes the average node value. $\epsilon$ denotes a small constant for numerical stability. $C$ denotes the exploration coefficient. $N_{p(i)}$ denotes the visit count of the parent node. Path reflections and global error memory jointly determine $H_i$. This paper instantiates the memory heuristic as:
\begin{equation}
H_i=\alpha Conf(M_{path,i}^{(k)})
-\beta Risk(M_{path,i}^{(k)},\mathcal{G}).
\end{equation}
For path records $m$, $Conf=\operatorname{clip}_{[0,1]}[0.5(1-\overline{sev})+0.3\min(\overline{\max(\Delta R,0)},1)+0.2\overline{I_{hint}}]$: it rewards lower mean severity, positive reward changes, and the presence of repair hints; it is zero for an empty path memory. $Risk$ is the maximum cosine similarity to the top-$t$ global records whose severity is at least 0.7, or zero if none exist. Thus, $\alpha$ and $\beta$ control a repair-confidence bonus and a high-risk similarity penalty. In all main runs, $C=1.4$, $\alpha=0.4$, and $\beta=0.6$; these are fixed design choices rather than theoretically optimal values.

To avoid repeated local edits around a wrong trajectory, \remcts{} uses two kinds of memory: path memory and global memory. The path memory $M_{path,i}^{(k)}$ stores the most recent $k$ structured reflections on the current branch. Each reflection contains the error type, failure symptom, root cause, repair hint, and severity. During expansion, the model only reads the recent failure trajectory. This controls context length and keeps the repair direction coherent.

The task-local global memory repository $\mathcal{G}$ stores error experience across branches of the current task and is reinitialized for every benchmark item. It is represented as:
\begin{align}
\mathcal{G}
&=\{(M_g,\mathbf{e}_g,Meta_g)\}_{g=1}^{|\mathcal{G}|},\\
\mathbf{e}_g&=Embed(M_g).
\end{align}
Here $M_g$ denotes a reflection summary. $\mathbf{e}_g$ denotes its embedding vector. $Meta_g$ denotes metadata that stores the source task, node id, error type, reward change, and risk level. We instantiate $Embed(\cdot)$ as a deterministic 128-dimensional hashed token-count vector and rank memories by cosine similarity. Expansion retrieves the top $t=2$ records without a hard similarity threshold. When a candidate is not accepted, the system writes its reflection summary, vector representation, and metadata into $\mathcal{G}$. The query uses the task, candidate code, and execution feedback. Retrieved memory enters the expansion prompt and semantic heuristic, but never deletes a branch; later execution evidence can override a misleading retrieval through numerical returns. This is a soft search signal, and the paired harm/benefit analysis in Section~\ref{sec:ablation} shows that it is useful in some harder settings rather than uniformly beneficial.

\begin{figure*}[t]
    \centering
    \includegraphics[width=\linewidth]{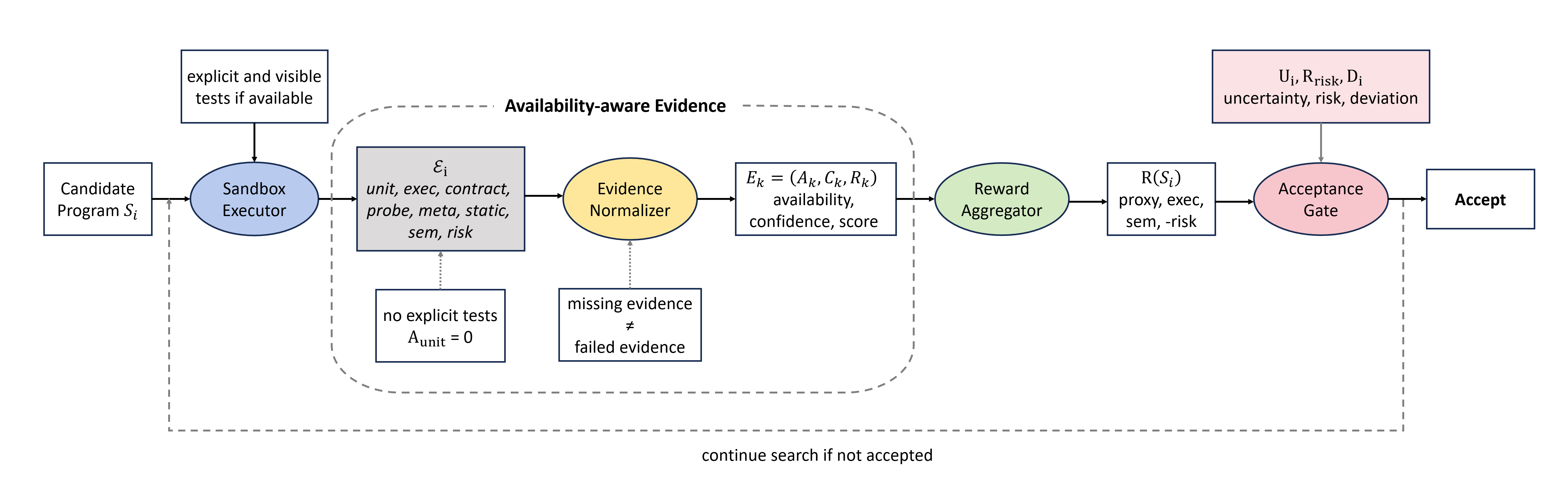}
    \caption{Evidence-driven reward and acceptance mechanism. A candidate program is first evaluated in the sandbox and converted into availability-aware evidence. When explicit tests are absent, unit-test evidence is marked unavailable ($A_{unit}=0$) rather than failed. The reward aggregator combines proxy correctness, execution reachability, semantic review, and risk. The acceptance gate further penalizes uncertainty, risk, and code-length deviation; rejected candidates return to search.}
    \label{fig:evidence-reward}
\end{figure*}

\subsection{Evidence-Driven Reward and Acceptance Mechanism}

This section defines an evidence-driven reward for code generation. \remcts{} evaluates each candidate with executable, interface, behavioral, semantic, and risk evidence, instead of relying on a single LLM score or a binary test result. Figure~\ref{fig:evidence-reward} summarizes how these signals are collected, aggregated into rewards, and used for final acceptance.

After a candidate program is generated, the sandbox executes it and collects weighted evidence:
\begin{equation}
\begin{split}
\mathcal{E}_i=\{&E_{unit},E_{exec},E_{contract},E_{probe},\\
&E_{meta},E_{static},E_{sem},E_{risk}\}.
\end{split}
\end{equation}
Here $\mathcal{E}_i$ denotes the evidence set collected for candidate program $S_i$. For each evidence item $E_k=(A_k,C_k,R_k)$, $A_k$ marks availability, $C_k$ is confidence, and $R_k$ is a normalized score. Unit tests provide the strongest signal when available; execution, contract, and static checks verify runnability and interface consistency; probe and meta evidence provide lightweight behavioral constraints in \notests{}; semantic review and risk checks support final filtering. Unavailable evidence is excluded from reward aggregation, so missing tests are not treated as failed tests. Contract evidence comes from AST rules for the required entry point and simple signature properties. In \notests{}, four LLM-generated weak assertions are derived from the task text alone, sanitized to prohibit imports, randomness, and file, network, or system access, and then executed as probes. Appendix Table~\ref{tab:appendix-evidence-main} gives the full evidence list and Section~\ref{app:implementation} gives implementation details.

In \notests{}, the final node reward is:
\begin{align}
R(S_i)
&=
\frac{
\sum_{g\in\mathcal{J}} \delta_g\omega_gA_gC_gR_g(S_i)
}{
\sum_{g\in\mathcal{J}} \omega_gA_gC_g+\epsilon
},\\
\mathcal{J}&=\{p,e,s,r\},\\
\delta_g&=
\begin{cases}
1, & g\in\{p,e,s\},\\
-1, & g=r.
\end{cases}
\end{align}
where $p,e,s,r$ index proxy correctness, execution reachability, semantic review, and risk, respectively. $A_g$ and $C_g$ are group availability and confidence, $\omega_g$ is the group weight, and $\delta_g$ assigns a negative sign only to risk. The \notests{} weights are $(\omega_p,\omega_e,\omega_s,\omega_r)=(0.4,0.2,0.3,0.1)$. In \known{}, the implementation instead uses $R_{known}=0.6R_{unit}^{obs}+0.2R_{exec}+0.2R_{sem}$ and accepts a candidate only after all available visible tests pass, subject to the search-level stopping rules. For risk-aware no-test acceptance, the system further computes:
\begin{align}
\mathrm{Score}_{acc}(S_i)=&\ R(S_i)-\gamma_1U_i\notag\\
&-\gamma_2R_{risk}(S_i)-\gamma_3D_i,
\end{align}
Here $U_i$ penalizes low-confidence evidence, $R_{risk}$ penalizes unsafe behavior, and $D_i$ is a code-length deviation penalty. If $\mathcal{Q}_i$ contains the available contract, static, meta, and probe items plus semantic and risk evidence, then $U_i=1-|\mathcal{Q}_i|^{-1}\sum_{q\in\mathcal{Q}_i}C_q$. The implementation sets $D_i$ to 0, 0.15, or 0.35 for at most 80, 81--200, or more than 200 nonblank lines. In \notests{}, acceptance requires $\mathrm{Score}_{acc}$, proxy, execution, and semantic thresholds to be met and risk to stay below its threshold. The gate is evaluated for every generated sibling; with complete-branch-before-accept enabled, the best accepted sibling is returned only after the branch is complete and the minimum iteration is reached.
\paragraph{Reward Component Details.}
We next define the item-level rewards used to compute proxy correctness, execution reliability, semantic review, and risk-aware acceptance.

The observed reward from explicit unit tests is:
\begin{equation}
R_{unit}^{obs}(S_i)=\frac{1}{m}\sum_{j=1}^{m}p_j(S_i),\quad m>0,
\end{equation}
where $m$ is the number of visible unit tests and $p_j(S_i)$ indicates whether $S_i$ passes the $j$-th test. If $m=0$, the method sets $A_{unit}=0$, so missing unit-test evidence is ignored rather than counted as failure.

The integrated proxy-correctness reward is:
\begin{equation}
\mathcal{K}_{proxy}=
\{\mathrm{contract},\mathrm{probe},
\mathrm{meta},\mathrm{static}\},
\end{equation}
\begin{equation}
R_{proxy}(S_i)=
\frac{\sum_{k\in\mathcal{K}_{proxy}} A_kC_k\eta_kR_k(S_i)}
{\sum_{k\in\mathcal{K}_{proxy}} A_kC_k\eta_k+\epsilon},
\end{equation}
where the fixed item weights for contract, probe, meta, and static evidence are $(0.35,0.25,0.15,0.25)$. Explicit unit tests are handled by $R_{unit}^{obs}$ in \known{} and remain unavailable in \notests{}.

Execution reward uses four observable indicators:
\begin{equation}
R_{exec}(S_i)=0.3(I_c+I_r)+0.2(I_t+I_o),
\end{equation}
where $I_c,I_r,I_t,I_o$ mark successful compilation, execution, the presence of executable tests, and nonempty output or successful execution, respectively. In \notests{}, let $Q_{sem}$ combine the reviewer's requirement coverage, API match, context compatibility, and maintainability scores with weights $(0.35,0.25,0.20,0.20)$. Given reviewer confidence $C_{sem}$ and semantic risk $q_{srisk}$, the implemented score is
\begin{equation}
R_{sem}=C_{sem}Q_{sem}(1-q_{srisk}).
\end{equation}
The separate static risk scan sets $R_{risk}=\operatorname{clip}(0.2n_{danger}+0.15I_{>200},0,1)$, where $n_{danger}$ counts prohibited imports or calls; a syntax error receives risk 0.25 because a full scan is impossible. Semantic review is therefore confidence- and risk-weighted rather than treated as a correctness oracle.

\section{Results}

\subsection{Experimental Setup}

\paragraph{Datasets.}
The main study uses HumanEval \citep{chen2021evaluating} and MBPP-Sanitized \citep{austin2021program}, which contain 164 and 427 function-level Python tasks. The protocol separates search feedback from final hidden evaluation. Visible feedback is derived from HumanEval prompt doctests and the first assertion of each MBPP-Sanitized task. We additionally run a limited HumanEval-X C++ pilot \citep{zheng2023codegeex} on the first 30 tasks in benchmark order, without task selection. File names, hashes, derivation details, and the pilot protocol are reported in Appendix~\ref{app:datasets}.

\paragraph{Models and Methods.}
This paper mainly focuses on open-weight non-GPT code models. We evaluate four dense Qwen3 models, Qwen3-1.7B, Qwen3-8B, Qwen3-14B, and Qwen3-32B \citep{Qwen3technicalreport}\footnote{Qwen3 model collection: \url{https://huggingface.co/collections/Qwen/Qwen3}}, together with DeepSeek-V4-Flash \citep{deepseekai2026deepseekv4} \footnote{DeepSeek-V4-Flash model: \url{https://huggingface.co/deepseek-ai/DeepSeek-V4-Flash}}. The main methods are Direct \notests{}, \remcts{} \notests{}, and \remcts{} \known{}. Supplementary baselines include Best-of-8, Reflexion-style, pure-MCTS, and MCTS-style, which test repeated sampling, single-trajectory reflection, tree search, and LLM semantic review. Budget-up controls use Best-of-32 and Reflexion-16 on representative MBPP \notests{} settings. We also report two component ablations that remove global memory retrieval or path reflection memory. Appendix~\ref{app:params} describes model invocation, decoding settings, and budget conventions.

\paragraph{Search Parameters and Metrics.}
All main \remcts{} experiments use a fixed search budget: 8 iterations, branch factor 5, maximum depth 8, minimum acceptance iteration 3, and complete-branch-before-accept enabled. Experiments use two search-feedback configurations: \known{} uses visible tests during search, while \notests{} removes explicit tests and relies on proxy evidence. Direct uses one generation, whereas \remcts{} and tree-search baselines use extra test-time compute. We report hidden solve rate as the main metric, exact two-sided McNemar tests for paired outcomes, Wilson 95\% intervals for solve rates, and token, call, wall-clock, and estimated monetary costs where logs are comparable. Appendix~\ref{app:params} gives the definitions and pricing assumptions.

\paragraph{Controlled-comparison scope.}
Our goal is to isolate the inference-time search mechanism, not to claim a leaderboard comparison against systems with different base models, fine-tuning data, generated or private tests, tools, or evaluation harnesses. The controlled baselines therefore preserve the base model, hidden-test evaluator, feedback access, and, for MCTS-style, the tree-search budget while replacing the search strategy or disabling ReMCTS components.

\subsection{Main Results}

The main result is that \remcts{} is substantially more effective when reliable feedback is available. Because the headline comparison changes both feedback access and test-time compute, we treat it as mechanism validation rather than attributing the entire gap to search or memory. Table~\ref{tab:main-results-short} reports hidden solve rates on HumanEval and MBPP-Sanitized. The \known{} setting is a test-guided mechanism validation: visible feedback can be used during search, while final scoring still uses only hidden tests. Therefore, this setting is not equivalent to standard pass@1. Under this protocol, \remcts{} \known{} improves over Direct \notests{} in 8 out of 10 model--dataset pairs, ties in 1 pair, and slightly decreases in 1 pair. It also improves over \remcts{} \notests{} in all pairs, showing that reliable feedback is more stable than proxy-only evidence.

\begin{table*}[t]
    \centering
    \caption{Hidden solve rates on HumanEval and MBPP-Sanitized. Parentheses in the two \remcts{} columns report percentage-point differences from Direct \notests{}. TG--\notests{} is the difference between \remcts{} \known{} and \remcts{} \notests{}. \known{} results are used for mechanism validation and are not equivalent to standard pass@1.}
    \small
    \begin{tabular}{@{}llcccc@{}}
        \toprule
        Model & Dataset & Direct \notests{} & \remcts{} \notests{} & \remcts{} \known{} & TG--\notests{} (pp) \\
        \midrule
        \texttt{Qwen3-1.7B} & HumanEval & 60.37\% & 69.51\% (+9.14) & 83.54\% (+23.17) & +14.03 \\
        \texttt{Qwen3-1.7B} & MBPP & 53.40\% & 51.29\% (-2.11) & 63.47\% (+10.07) & +12.18 \\
        \texttt{Qwen3-8B} & HumanEval & 79.88\% & 86.59\% (+6.71) & 93.90\% (+14.02) & +7.31 \\
        \texttt{Qwen3-8B} & MBPP & 64.17\% & 66.74\% (+2.57) & 74.24\% (+10.07) & +7.50 \\
        \texttt{Qwen3-14B} & HumanEval & 82.93\% & 78.66\% (-4.27) & 82.93\% (+0.00) & +4.27 \\
        \texttt{Qwen3-14B} & MBPP & 66.98\% & 60.42\% (-6.56) & 65.57\% (-1.41) & +5.15 \\
        \texttt{Qwen3-32B} & HumanEval & 89.63\% & 89.63\% (+0.00) & 95.12\% (+5.49) & +5.49 \\
        \texttt{Qwen3-32B} & MBPP & 71.90\% & 70.49\% (-1.41) & 77.99\% (+6.09) & +7.50 \\
        \texttt{DeepSeek-V4-Flash} & HumanEval & 80.49\% & 95.12\% (+14.63) & 98.78\% (+18.29) & +3.66 \\
        \texttt{DeepSeek-V4-Flash} & MBPP & 72.60\% & 72.83\% (+0.23) & 81.03\% (+8.43) & +8.20 \\
        \bottomrule
    \end{tabular}
    \label{tab:main-results-short}
    \vspace{-8pt}
\end{table*}
\paragraph{Paired statistical evidence.}
Pooling the 2,955 model--task pairs, Direct \notests{} solves 2,050 (69.37\%) and \remcts{} \known{} solves 2,292 (77.56\%, Wilson 95\% CI 76.02--79.03). There are 364 \remcts{}-only and 122 Direct-only successes (two-sided exact McNemar $p=5.3\times10^{-29}$). Eight individual settings improve significantly at $p<0.05$; Qwen3-14B HumanEval ties and Qwen3-14B MBPP decreases non-significantly. Appendix Table~\ref{tab:appendix-main-significance} reports all paired counts, confidence intervals, and $p$-values. The pooled test treats each model--task output as a paired observation; the individual tests avoid interpreting the pooled result as evidence about variation across models.

\paragraph{Model Capacity and Proxy Evidence.}
Two secondary patterns qualify the main gain: improvements depend on model capacity, and proxy-only evidence can still miss hidden semantics. The gaps in Table~\ref{tab:main-results-short}, search-tree statistics in Appendix~\ref{app:tree}, and trajectories in Appendix Table~\ref{tab:case-analysis} show that weaker models often benefit from execution evidence. Stronger models such as Qwen3-32B mainly recover residual errors. The Qwen3-14B drops and some MBPP \notests{} cases expose the proxy-evidence boundary: plausible candidates can pass weak checks while still failing hidden semantics.

\subsection{Cross-Language Pilot}
\label{sec:cpp-pilot}
To test executor portability, we ran a HumanEval-X C++ pilot on the first 30 tasks in published order, without filtering. Search used only the public \texttt{example\_test}; the official \texttt{test} field was held out. With a C++17 executor and a smaller 3-iteration, branch-3 budget, \remcts{} improves hidden solve rate from 73.33 to 90.00 on Qwen3-8B, 83.33 to 96.67 on Qwen3-32B, and 60.00 to 96.67 on DeepSeek-V4-Flash. Pooling the 90 paired outputs gives 21 \remcts{}-only and 1 Direct-only success ($p=1.10\times10^{-5}$). Appendix Table~\ref{tab:cpp-pilot} reports the protocol, tokens, and costs. This limited pilot establishes compiler-backed compatibility, not broad multilingual capability.

\subsection{Budget-Up and Cost Analysis}
\label{sec:budget}
To test whether the main gains are explained by generating more candidates, we increased the budgets of the two strongest simple controls on representative MBPP \notests{} runs. Best-of-32 generates 32 independent candidates, and Reflexion-16 permits up to 16 actor--evaluator--reflection rounds. These runs are not strict wall-clock matches; they test whether increasing sampling or repair budget closes the accuracy gap.

\begin{table*}[t]
    \centering
    \caption{Budget-up baselines on MBPP \notests{}. Wall-clock is total elapsed time over 427 tasks; tokens and calls are per task. Hidden tests are used only for final evaluation.}
    \label{tab:budget-up}
    \small
    \setlength{\tabcolsep}{3.4pt}
    \begin{tabular}{@{}llrrrrr@{}}
        \toprule
        Model & Method & Hidden solve & Tok/task & Calls/task & Wall (s) & Cost (CNY) \\
        \midrule
        \texttt{Qwen3-8B} & ReMCTS & 66.74 & 10,295 & 19.7 & 1,980 & 3.72 \\
        \texttt{Qwen3-8B} & Best-of-32 & 59.95 & 8,290 & 44.7 & 1,502 & 3.18 \\
        \texttt{Qwen3-8B} & Reflexion-16 & 62.30 & 15,777 & 17.1 & 903 & 4.54 \\
        \texttt{Qwen3-32B} & ReMCTS & 70.49 & 9,735 & 19.7 & 3,053 & 13.15 \\
        \texttt{Qwen3-32B} & Best-of-32 & 64.17 & 8,600 & 33.0 & 2,232 & 13.79 \\
        \texttt{Qwen3-32B} & Reflexion-16 & 69.09 & 11,072 & 12.4 & 1,147 & 12.94 \\
        \bottomrule
    \end{tabular}
\end{table*}

Best-of-32 remains 6.79 and 6.32 percentage points below ReMCTS for Qwen3-8B and Qwen3-32B, respectively. Reflexion-16 remains 4.44 and 1.40 points below; notably, Qwen3-8B Reflexion-16 uses more logged tokens than ReMCTS (15.8k vs. 10.3k) but is less accurate. ReMCTS therefore trades additional sequential search overhead for robustness and should not be interpreted as the cheapest or fastest method. The cost conclusion is setting-dependent: Reflexion-style is more cost-effective on known-feedback MBPP, where visible assertions are strong, whereas ReMCTS is more robust to incomplete feedback and outperforms the controlled memory-free tree-search baseline in the comparisons below.

\paragraph{Supplementary Baseline Setup.}
The supplementary baselines test whether repeated sampling, single-path reflection, or tree search alone explains the gains. We compare Best-of-8, Reflexion-style, pure-MCTS, and MCTS-style on the same three models, without hidden tests for search or selection. MCTS-style is the closest memory-free tree-search baseline: it adds LLM semantic review but disables path memory, global memory, semantic UCB, and LLM probes. Full results, acceptance rates, Oracle Pool upper bounds, and configuration details are in Appendix Tables~\ref{tab:sampling-reflexion-baselines} and~\ref{tab:appendix-known-feedback-baselines}.

\paragraph{Sampling and Reflection Baselines.}
In matched no-test runs, Best-of-8 trails \remcts{} on both Qwen MBPP settings, while Reflexion-style trails in five of six pairs (except Qwen3-32B HumanEval). Best-of-8 also trails \remcts{} \known{} throughout, but that cross-feedback comparison is descriptive. The offline Oracle Pool sometimes contains a correct candidate that the baseline fails to select.

\paragraph{Tree-Search Baselines.}
Tree search alone is also insufficient: in \notests{}, pure-MCTS and MCTS-style remain below full \remcts{} on all three MBPP settings. In \known{}, full \remcts{} beats MCTS-style in all six pairs by 5.07 points on average. The strongest paired gains are Qwen3-8B HumanEval (13 vs. 1, $p=0.00183$), Qwen3-8B MBPP (33 vs. 1, $p=4.07\times10^{-9}$), Qwen3-32B MBPP (30 vs. 1, $p=2.98\times10^{-8}$), and DeepSeek MBPP (27 vs. 5, $p=1.13\times10^{-4}$); larger-model HumanEval comparisons are near ceiling. Appendix Tables~\ref{tab:appendix-mcts-significance} and~\ref{tab:appendix-token-comparison} give all paired and token results.

The \known{}-feedback Reflexion-style results show that visible tests can make single-path repair highly competitive on short assertion-style tasks. Full \remcts{} remains higher on HumanEval for all three models, but Reflexion-style is stronger on MBPP, reaching 81.97\%, 86.89\%, and 94.15\% on Qwen3-8B, Qwen3-32B, and DeepSeek-V4-Flash. Thus, \remcts{}'s advantage is not simply that reflection helps; multi-branch search and reusable memory mainly improve robustness when feedback is incomplete, noisy, or hard to exploit through one trajectory.

\subsection{Ablation Results}
\label{sec:ablation}

\paragraph{Memory Ablations.}
Table~\ref{tab:memory-ablation-main} reports the core component results. Removing global retrieval lowers hidden solve rate by 0.97 points on average over twelve settings, with the clearest paired effects on Qwen3-8B MBPP-\known{} (23 full-only vs. 3 no-global-only, $p=8.8\times10^{-5}$) and Qwen3-32B MBPP-\known{} (17 vs. 3, $p=0.00258$). The effect is noisy on DeepSeek MBPP-\known{} (12 vs. 10, $p=0.83$), while path memory has a mixed 5-win/4-tie/3-loss pattern. Appendix Table~\ref{tab:appendix-memory-ablation} adds acceptance and tree statistics.

\begin{table*}[!t]
    \centering
    \caption{Memory ablations under final hidden evaluation (\%). Full denotes complete \remcts{}; the other columns disable one memory component.}
    \label{tab:memory-ablation-main}
    \scriptsize
    \setlength{\tabcolsep}{5.2pt}
    \renewcommand{\arraystretch}{0.88}
    \begin{tabular}{@{}lllrrr@{}}
        \toprule
        Model & Dataset & Setting & Full & W/O global & W/O path \\
        \midrule
        \texttt{DeepSeek-V4-Flash} & HumanEval & \known{}   & 98.78 & 99.39 & 99.39 \\
        \texttt{DeepSeek-V4-Flash} & HumanEval & \notests{} & 95.12 & 94.51 & 94.51 \\
        \texttt{DeepSeek-V4-Flash} & MBPP      & \known{}   & 81.03 & 80.56 & 80.09 \\
        \texttt{DeepSeek-V4-Flash} & MBPP      & \notests{} & 72.83 & 73.07 & 72.83 \\
        \midrule
        \texttt{Qwen3-8B} & HumanEval & \known{}   & 93.90 & 92.07 & 95.12 \\
        \texttt{Qwen3-8B} & HumanEval & \notests{} & 86.59 & 85.37 & 86.59 \\
        \texttt{Qwen3-8B} & MBPP      & \known{}   & 74.24 & 69.56 & 73.54 \\
        \texttt{Qwen3-8B} & MBPP      & \notests{} & 66.74 & 66.04 & 66.51 \\
        \midrule
        \texttt{Qwen3-32B} & HumanEval & \known{}   & 95.12 & 95.12 & 95.12 \\
        \texttt{Qwen3-32B} & HumanEval & \notests{} & 89.63 & 90.85 & 89.63 \\
        \texttt{Qwen3-32B} & MBPP      & \known{}   & 77.99 & 74.71 & 78.69 \\
        \texttt{Qwen3-32B} & MBPP      & \notests{} & 70.49 & 69.56 & 70.02 \\
        \bottomrule
    \end{tabular}
    \vspace{-5pt}
\end{table*}

\paragraph{Search-Tree Behavior.}
Search-tree statistics indicate that \remcts{} usually performs shallow multi-branch exploration rather than long repair chains. Appendix Table~\ref{tab:appendix-tree-stats} gives the full per-model values. The average tree size ranges from 3.389 to 11.593 nodes, the median mostly falls between 3 and 8 nodes, and the average maximum depth is about 1.23--1.61. A smaller actual tree does not mean the parameters are inconsistent: \texttt{branch\_factor}=5 is only an upper bound, and parsing failures, duplicate code, cross-node deduplication, and early acceptance all reduce the number of effective children.

\paragraph{Minimum Acceptance Iteration.}
The minimum-acceptance ablation further shows that no-test deployment needs stronger uncertainty control. Increasing \texttt{min\_accept\_iterations} from 1 to 3 improves hidden solve rate in 7 of 10 \notests{} combinations, with an average gain of 1.55 points; full scores and search-scale changes are reported in Appendix Tables~\ref{tab:appendix-min-accept-score} and~\ref{tab:appendix-min-accept}.

\paragraph{Qualitative Error Analysis.}
Representative trajectories (Appendix Table~\ref{tab:case-analysis}) make the aggregate patterns concrete. In known-feedback HumanEval, Qwen3-1.7B recovers an \texttt{intersperse} failure by exploring siblings after early candidates fail visible tests; the selected branch changes delimiter placement instead of repeatedly patching the same implementation. For Qwen3-8B \texttt{remove\_duplicates}, path memory records a semantic confusion and a missing import before a later branch passes hidden tests. For Qwen3-32B \texttt{starts\_one\_ends}, the search repairs an inclusion--exclusion error and the $n=1$ boundary case. These examples suggest that weaker models benefit from execution-guided filtering, whereas stronger models mainly use search to recover residual edge cases.

In \notests{}, the failure case is different. A Qwen3-14B MBPP candidate is accepted after contract and static checks pass, despite a partially failing weak probe and over-optimistic semantic review; hidden tests later reject it. This illustrates why proxy-only gains are less stable and why delaying acceptance improves seven of ten combinations. More generally, automatic acceptance is a risk-controlled stopping rule, not a correctness oracle: hidden outcomes are never exposed during search, and offline hidden evaluation is used only to audit the returned candidate. The paired statistics therefore measure end-to-end success, while trajectories explain plausible mechanisms rather than establish causal mediation.

\section{Conclusions}

This paper presents \remcts{}, an execution-grounded, memory-augmented, LLM-guided MCTS-style framework for inference-time code search. On HumanEval and MBPP-Sanitized, visible feedback gives the strongest hidden-test gains; \notests{} is positive on average but unstable. Paired tests and budget-up controls show that the gains are not explained by more samples alone. Global memory contributes more consistently than path memory. A 30-task C++ pilot suggests compiler-backed transfer, not broad multilingual or repository-level capability. Future work should strengthen proxy feedback and uncertainty calibration.

\clearpage
\section*{Limitations}

Evaluation is limited to function-level benchmarks: \known{} is test-guided mechanism validation rather than standard pass@1, while the 30-task C++ pilot cannot establish broad multilingual or repository-level capability. In both settings, hidden tests are used only for offline scoring and never guide search. Proxy signals can miss hidden boundary cases, the executor is not security-hardened, and sequential calls add cost. We do not evaluate workspace patches, cross-file dependencies, environment configuration, test selection, or long-horizon debugging, nor isolate every heuristic.

Cost comparisons should also be interpreted cautiously. API latency, batching, server load, and model-version changes can affect wall-clock time and token accounting. We therefore report tokens, calls, and elapsed time together as within-study diagnostics, rather than claiming deployment-wide efficiency.

\section*{Acknowledgments}

This work is supported by the Intelligent Computing Center of Shenzhen University.
During the preparation of this manuscript and its accompanying code, the authors
used OpenAI ChatGPT/Codex for language editing, \LaTeX{} organization, code and
documentation assistance, and reference cross-checking. All AI-generated
suggestions were reviewed and verified by the authors, who take full
responsibility for the final content.

\clearpage
\bibliography{ref}

\clearpage
\appendix

\section{\remcts{} Pseudocode}
\label{app:pseudocode}

Algorithm~\ref{alg:remcts} summarizes the full search procedure. In \known{}, \textsc{EvidenceGate} requires all visible tests to pass; in \notests{}, unit tests are unavailable and the gate applies the proxy, execution, semantic, risk, and acceptance-score thresholds. The algorithm evaluates the complete generated branch before returning its highest-reward accepted sibling, and does not stop before the minimum iteration.

\begin{algorithm}[H]
    \caption{\remcts{} search procedure}
    \label{alg:remcts}
    \small
    \begin{algorithmic}[1]
        \Require Task $P$, optional visible tests $\mathcal{T}_{vis}$, model $f_\theta$
        \Require Budget $B$, branch factor $b$, max depth $d$, memory window $k$, minimum iteration $i_{min}$
        \Ensure Final candidate program $S^\star$
        \State Initialize root node $v_0$ and global memory $\mathcal{G}\leftarrow\varnothing$
        \State $S^\star,S_{acc}\leftarrow\varnothing$; $r^\star,r_{acc}\leftarrow-\infty$
        \For{$iter=1$ to $B$}
            \State $v \leftarrow \Call{Select}{v_0,UCB_M,d}$
            \State $M_{path}\leftarrow \Call{PathMemory}{v,k}$
            \State $M_{global}\leftarrow \Call{Retrieve}{\mathcal{G},P,S_v,\Phi_v}$
            \State $X\leftarrow \Call{BuildPrompt}{P,S_v,\Phi_v,M_{path},M_{global}}$
            \State $\mathcal{C}\leftarrow \Call{Generate}{f_\theta,X,b}$
            \ForAll{$S_i\in\mathcal{C}$}
                \State $\mathcal{E}_i\leftarrow \Call{CollectEvidence}{P,S_i,\mathcal{T}_{vis}}$
                \If{$\mathcal{T}_{vis}=\varnothing$}
                    \State Set $A_{unit}\leftarrow 0$
                \EndIf
                \State $R_i\leftarrow \Call{EvidenceReward}{\mathcal{E}_i}$
                \State $g_i\leftarrow \Call{EvidenceGate}{S_i,\mathcal{E}_i,R_i}$
                \State $M_i\leftarrow\varnothing$
                \If{not $g_i$}
                    \State $M_i\leftarrow \Call{Reflect}{P,S_i,\mathcal{E}_i,M_{path},M_{global}}$
                \EndIf
                \State $u\leftarrow \Call{AddChild}{v,S_i,R_i,\mathcal{E}_i,M_i}$
                \State \Call{Backpropagate}{$u,R_i$}
                \If{\Call{Storeable}{$M_i,\mathcal{E}_i$}}
                    \State $\mathcal{G}\leftarrow \mathcal{G}\cup\{(M_i,\Call{Embed}{M_i},Meta_i)\}$
                \EndIf
                \If{$R_i>r^\star$}
                    \State $S^\star\leftarrow S_i$, $r^\star\leftarrow R_i$
                \EndIf
                \If{$g_i$ and $R_i>r_{acc}$}
                    \State $S_{acc}\leftarrow S_i$, $r_{acc}\leftarrow R_i$
                \EndIf
            \EndFor
            \If{$S_{acc}\neq\varnothing$ and $iter\geq i_{min}$}
                \State \Return $S_{acc}$
            \EndIf
        \EndFor
        \State \Return $S^\star$
    \end{algorithmic}
\end{algorithm}

\section{Reward and Evidence Details}
\label{app:reward-evidence-details}

Symbols in the method formulas fall into three categories. The first category is predefined hyperparameters, including the search budget, $C$, memory window $k$, global retrieval count $t$, semantic heuristic weights $\alpha,\beta$, evidence and reward weights $\eta_k,\omega_g$, acceptance penalties $\gamma_j$, and thresholds $\tau_*$. The second category is runtime evidence quantities, including $A_k,C_k,R_k$ and reward components derived from them. The third category is rule- or LLM-assisted scores, including $H_i$, semantic reviews, risk scans, and the deviation penalty $D_i$. Hyperparameters are fixed before evaluation and are not learned from tasks. No model or reward parameters are fitted; search-time state updates consist of tree structure and node statistics, current-best state, evidence metadata, and task-local reflection memory.

Table~\ref{tab:appendix-symbol-types} summarizes these symbol categories and fixed hyperparameter settings. Table~\ref{tab:appendix-evidence-main} lists the evidence items used to evaluate candidate programs. If an evidence item is unavailable, $A_k=0$, and the missing item is ignored by the reward denominator.

\begin{table*}[t]
    \centering
    \caption{Symbol categories and hyperparameter settings.}
    \small
    \begin{tabularx}{\textwidth}{@{}lXX@{}}
        \toprule
        Symbol & Type & Definition \\
        \midrule
        $C,k,t,\alpha,\beta$ & Search and semantic-heuristic hyperparameters & $C=1.4$, $k=3$, $t=2$; \remcts{} uses $\alpha=0.4,\beta=0.6$, while pure-MCTS/MCTS-style ablations set the memory heuristic to zero. \\
        $\omega_g,\gamma_j,\tau_*$ & Reward and acceptance hyperparameters & \known{} uses unit/exec/semantic weights $(0.6,0.2,0.2)$; \notests{} uses proxy/exec/semantic/risk weights $(0.4,0.2,0.3,0.1)$. The \notests{} acceptance penalties are $\gamma=(0.10,0.35,0.10)$, and its thresholds are $\tau_{proxy}=0.50$, $\tau_{exec}=0.60$, $\tau_{sem}=0.45$, $\tau_{risk}=0.25$, and $\tau_{accept}=0.62$. \\
        $\eta_k$ & Proxy-evidence weights & Contract/probe/meta/static weights are $(0.35,0.25,0.15,0.25)$; fixed before evaluation. \\
        $A_k,C_k,R_k,I_{\cdot},R_{\cdot},U_i$ & Runtime evidence quantities & Produced by visible tests, sandbox execution, contract checks, probes, property checks, static checks, risk scans, and LLM reviews; $U_i$ is computed from the confidence of available evidence groups. \\
        $H_i,D_i,R_{sem}$ & Rule- or LLM-assisted scores & $H_i$ combines path-reflection confidence and high-severity global-memory similarity; $D_i$ depends on nonblank code length; $R_{sem}$ is a confidence- and risk-weighted LLM review. \\
        $\epsilon$ & Numerical stability term & Used only to avoid division by zero; not a tunable experimental parameter. \\
        \bottomrule
    \end{tabularx}
    \label{tab:appendix-symbol-types}
\end{table*}

\begin{table*}[t]
    \centering
    \caption{Evidence items used in reward computation.}
    \small
    \begin{tabularx}{\textwidth}{@{}lXX@{}}
        \toprule
        Evidence & Meaning & Role \\
        \midrule
        Unit & Visible unit tests or assertions & Provides the highest-confidence functional feedback in \known{} \\
        Exec & Compilation, loading, entry call, timeout, and crash status & Checks whether the candidate is runnable and callable \\
        Contract & Function name, argument count, return form, or schema & Prevents interface mismatches and entry-point errors \\
        Probe & Four LLM-generated weak assertions from task text; fixed confidence 0.45 & Provides low-cost behavioral signals without explicit tests after sanitizer checks \\
        Meta & Properties or metamorphic relations extracted from the requirement; fixed confidence 0.35 & Checks task properties such as sorting, length, idempotence, and reversal \\
        Static & Compilation/execution status; fixed confidence 0.80 & Filters obviously non-runnable code \\
        Semantic & LLM semantic review & Estimates consistency between the candidate approach and task description \\
        Risk & Dangerous system calls, unauthorized file access, abnormal complexity, etc. & Provides risk penalties for production-style acceptance \\
        \bottomrule
    \end{tabularx}
    \label{tab:appendix-evidence-main}
\end{table*}

\section{Datasets and Input Settings}
\label{app:datasets}

Table~\ref{tab:appendix-datasets} lists the data files, task counts, and source summaries. Complete SHA-256 hashes, byte counts, and derivation relations are recorded in \texttt{data/dataset\_manifest.json}.

\begin{table*}[t]
    \centering
    \caption{Experimental data files and usage.}
    \label{tab:appendix-datasets}
    \small
    \begin{tabularx}{\textwidth}{@{}lrlX@{}}
        \toprule
        File & Tasks & Use & Source or Description \\
        \midrule
        \texttt{HumanEval\_feedback.jsonl} & 164 & \known{} input & Visible feedback parsed from doctest examples in HumanEval prompts; some tasks have no visible examples \\
        \texttt{HumanEval\_no\_tests.jsonl} & 164 & \notests{} input & Explicit tests removed; only the prompt and entry point are retained \\
        \texttt{HumanEval\_eval.jsonl} & 164 & hidden eval & Original HumanEval test fields retained only for final offline evaluation \\
        \texttt{mbpp\_sanitized\_feedback\_split.json} & 427 & \known{} input & Converted from Google Research \texttt{sanitized-mbpp.json}; only the first assertion per task is used as search feedback \\
        \texttt{mbpp\_sanitized\_no\_tests.json} & 427 & \notests{} input & Explicit assertions removed; only task description and entry point are retained \\
        \texttt{mbpp\_sanitized\_eval\_split.json} & 427 & hidden eval & Setup/imports and remaining assertions retained for final evaluation \\
        \texttt{HumanEval-X C++ first 30} & 30 & cross-language pilot & Public \texttt{example\_test} for search; official \texttt{test} retained only for final evaluation \\
        \bottomrule
\end{tabularx}
\end{table*}

The experiments use existing public function-level code-generation benchmarks,
including HumanEval, MBPP-Sanitized, and HumanEval-X, and do not collect human
or user-generated data. We checked the released task files for obvious direct
identifiers and offensive-content concerns. No personal identifiers are used in
our analyses, and the benchmark materials are retained and redistributed only
in accordance with their original source terms. No additional anonymization was
required.

\paragraph{Cross-Language Pilot.}
Table~\ref{tab:cpp-pilot} gives the full results for the first 30 HumanEval-X C++ tasks in published order. Search uses only \texttt{example\_test}; the official \texttt{test} field is held out for final evaluation. The C++17 runs use 3 iterations, branch factor 3, maximum depth 4, minimum acceptance iteration 2, and an 8-second execution timeout.

\begin{table*}[t]
    \centering
    \caption{HumanEval-X C++ pilot on the first 30 tasks. Official tests are used only for final hidden evaluation. Costs use the logged-token pricing assumptions of the Python experiments.}
    \label{tab:cpp-pilot}
    \small
    \setlength{\tabcolsep}{4pt}
    \begin{tabular}{@{}llrrrrr@{}}
        \toprule
        Model & Method & Tasks & Visible pass & Hidden solve & Tok/task & Cost (CNY) \\
        \midrule
        \texttt{Qwen3-8B} & Direct & 30 & 73.33 & 73.33 & 335 & 0.0094 \\
        \texttt{Qwen3-8B} & ReMCTS & 30 & 93.33 & 90.00 & 3,217 & 0.0818 \\
        \texttt{Qwen3-32B} & Direct & 30 & 83.33 & 83.33 & 337 & 0.0381 \\
        \texttt{Qwen3-32B} & ReMCTS & 30 & 96.67 & 96.67 & 3,245 & 0.3260 \\
        \texttt{DeepSeek-V4-Flash} & Direct & 30 & 60.00 & 60.00 & 330 & 0.0123 \\
        \texttt{DeepSeek-V4-Flash} & ReMCTS & 30 & 96.67 & 96.67 & 2,996 & 0.1063 \\
        \bottomrule
    \end{tabular}
\end{table*}

\section{Implementation and Reproducibility}
\label{app:implementation}

The complete release accompanying the camera-ready paper includes the search and evaluation framework, prompt templates, configuration files, dataset preprocessing and split scripts, the Python and C++ compiler-backed executors, hidden-evaluation scripts, anonymized per-task outputs, reflection/retrieval logs, and token/cost logs. API-dependent runs record the exact model alias, run date, decoding parameters, data hashes, and saved anonymized outputs so that hidden scores can be recomputed without re-querying a mutable hosted service. Secrets and provider credentials are excluded from the release.

The main Python runs use a 5-second subprocess timeout and the search settings in Table~\ref{tab:appendix-search-params}. The \notests{} gate uses $w_{proxy}/w_{exec}/w_{semantic}/w_{risk}=0.4/0.2/0.3/0.1$, thresholds $(\tau_{accept},\tau_{proxy},\tau_{exec},\tau_{sem},\tau_{risk})=(0.62,0.50,0.60,0.45,0.25)$, and penalties $(\gamma_{uncertainty},\gamma_{risk},\gamma_{deviation})=(0.10,0.35,0.10)$. Fixed evidence confidences are $C_{probe}=0.45$, $C_{contract}=0.75$, $C_{static}=0.80$, and $C_{meta}=0.35$. LLM calls use temperature 0.2 and at most 1024 output tokens. Global memory uses deterministic 128-dimensional hashed token-count embeddings and cosine similarity; it retrieves the top two memories without a hard similarity threshold. In \notests{}, probe generation uses four weak assertions from task text only, sanitized against imports, randomness, file/network/system calls, and executed in the subprocess sandbox.

For cost accounting, we use the logged prompt and completion tokens and the pricing assumptions recorded with each run. The Qwen experiments use CNY 0.3 per million input tokens and CNY 1.2 per million output tokens; DeepSeek reports use the provider-specific rates saved in the corresponding report. Wall-clock values are total elapsed time for the complete task set, not per-task latency, and parallel workers can make them non-additive. These details explain why cost and wall time should be read as operational accounting rather than a strict compute-matched comparison.

\section{Search Parameters and Evaluation Metrics}
\label{app:params}

Table~\ref{tab:appendix-model-settings} reports model invocation and decoding settings. All models are called through an OpenAI-compatible API. They are identified by the API aliases in our experiment records. API-served models are subject to server-side updates. Therefore, we fix the experimental protocol with aliases, run time, decoding parameters, and data hashes.

\begin{table*}[t]
    \centering
    \caption{Model invocation and decoding settings.}
    \label{tab:appendix-model-settings}
    \small
    \begin{tabularx}{\textwidth}{@{}lX@{}}
        \toprule
        Item & Setting \\
        \midrule
        Model alias & \texttt{Qwen3-1.7B}, \texttt{Qwen3-8B}, \texttt{Qwen3-14B}, \texttt{Qwen3-32B}, \texttt{DeepSeek-V4-Flash} \\
        Invocation & OpenAI-compatible API; main experiments were run in May 2026 and the budget-up/C++ pilots in July 2026 \\
        Direct, \remcts{}, and tree-search baselines & Code-generation temperature 0.2, maximum output length 1024 tokens, and API request timeout 90 seconds \\
        Best-of-8 & 8 independent candidates per task, generation temperature 0.7, and 5 self-generated assertion tests for selection \\
        Best-of-32 & 32 independent candidates per task with the same generation and proxy-selection protocol as Best-of-8 \\
        Reflexion-style & Up to 8 Actor--Evaluator--Self-Reflection rounds; Actor, Evaluator, and reflection temperatures are all 0.2 \\
        Reflexion-16 & Up to 16 rounds with the same Actor--Evaluator--Self-Reflection protocol \\
        Sandbox execution & Python timeout 5 seconds; C++17 pilot timeout 8 seconds; hidden tests are used only for final offline evaluation and Oracle Pool statistics \\
        \bottomrule
    \end{tabularx}
\end{table*}

Table~\ref{tab:appendix-search-params} gives the search budget used in the main experiments. To make \known{} and \notests{} comparable, they use the same maximum iterations, branch factor, maximum depth, and minimum acceptance iterations. Direct is a single-generation baseline. Best-of-8 generates exactly 8 candidates. Reflexion-style runs up to 8 single-path feedback rounds. \remcts{}, pure-MCTS, and MCTS-style use the same tree-search budget. If a tree has $n$ nodes, then the $n-1$ non-root nodes correspond to generated and executed program candidates. Semantic reviews, reflections, and LLM probes add auxiliary calls when they are enabled. Different API services do not expose perfectly consistent token accounting, so the main analysis uses average iterations and search-tree size as compute-budget proxies across models and methods. For the direct \remcts{}/MCTS-style comparison, Appendix Table~\ref{tab:appendix-token-comparison} additionally reports logged prompt-plus-completion token totals from the corresponding experiment reports.

\begin{table*}[t]
    \centering
    \caption{Search parameters for the main \remcts{} experiments.}
    \label{tab:appendix-search-params}
    \small
    \begin{tabularx}{\textwidth}{@{}>{\raggedright\arraybackslash}X>{\raggedright\arraybackslash}X@{}}
        \toprule
        Parameter & Value \\
        \midrule
        \texttt{max\_iterations} & 8 \\
        \texttt{branch\_factor} & 5 \\
        \texttt{max\_depth} & 8 \\
        \texttt{no\_improve\_patience} & 8 \\
        \texttt{min\_accept\_iterations} & 3 \\
        \texttt{complete\_branch\_before\_accept} & true \\
        Final metric & hidden solve rate \\
        \bottomrule
    \end{tabularx}
\end{table*}

Let $a_i\in\{0,1\}$ indicate whether the internal acceptor of a method automatically accepts task $i$. Let $h_i\in\{0,1\}$ indicate whether the final returned candidate passes hidden tests. Hidden Solve Rate is computed over all tasks as $\frac{1}{n}\sum_i h_i$. It does not depend on automatic acceptance. Accepted Rate is $\frac{1}{n}\sum_i a_i$. Acceptance Precision is $\frac{\sum_i a_i h_i}{\sum_i a_i+\epsilon}$. False Acceptance Rate is $\frac{\sum_i a_i(1-h_i)}{\sum_i a_i+\epsilon}$. For \remcts{}, pure-MCTS, and MCTS-style, $a_i$ comes from their acceptance thresholds. For Best-of-8 and Reflexion-style, $a_i$ means that self-generated tests or the evaluator accepted a candidate. If $a_i=0$, the system still returns the current best candidate for hidden evaluation. Thus, the task affects Hidden Solve Rate, but it is not included in the denominator of acceptance precision or false acceptance. Table~\ref{tab:appendix-metrics} summarizes these metric definitions.

\begin{table*}[t]
    \centering
    \caption{Definitions of evaluation metrics.}
    \label{tab:appendix-metrics}
    \small
    \begin{tabularx}{\textwidth}{@{}lXX@{}}
        \toprule
        Metric & Definition & Applicable Settings \\
        \midrule
        Hidden Solve Rate & Fraction of tasks whose final candidate passes hidden tests & All experiments \\
        Accepted Rate & Fraction of tasks whose candidate is automatically accepted & Methods/settings with an internal acceptor \\
        Acceptance Precision & Fraction of automatically accepted candidates that pass hidden tests & Methods/settings with an internal acceptor \\
        False Acceptance Rate & Fraction of automatically accepted candidates that fail hidden tests & Methods/settings with an internal acceptor \\
        Avg. Iterations & Average search iterations per task & Search efficiency analysis \\
        Avg. Tree Size & Average search-tree nodes per task & Search efficiency analysis \\
        Oracle Pool Hidden Rate & Fraction of tasks where at least one candidate in the pool passes hidden tests & Offline analysis for multi-candidate and tree-search methods \\
        \bottomrule
    \end{tabularx}
\end{table*}

\section{Supplementary Baseline Results}
\label{app:baselines}

Table~\ref{tab:sampling-reflexion-baselines} reports the full supplementary results for Best-of-8, Reflexion-style, pure-MCTS, and MCTS-style. Hidden tests are used only for final offline evaluation. They are never used for candidate selection.

\begin{table*}[t]
    \centering
    \caption{Supplementary baselines for Best-of-8, Reflexion-style, pure-MCTS, and MCTS-style. Accepted is the number of tasks accepted by the method's internal acceptor. It is not the number of final returned tasks. Hidden Solve is always computed with the final returned candidate for each task. Oracle Pool is the offline upper bound indicating whether at least one candidate in the pool passes hidden tests. Best-of-8 uses 8 independent candidates. Reflexion-style uses all attempts. pure-MCTS and MCTS-style use non-root nodes generated in the search tree. \texttt{DeepSeek} abbreviates DeepSeek-V4-Flash.}
    \label{tab:sampling-reflexion-baselines}
    \vspace{2pt}
    \begin{minipage}{0.98\textwidth}
    \footnotesize
    \textit{Parameter note:} Best-of-8 uses $n=8$ independent candidates. It uses a candidate-generation temperature of 0.7 and 5 self-generated assertion tests for selection. Reflexion-style runs up to 8 Actor--Evaluator--Self-Reflection rounds. Actor and reflection generation temperatures are both 0.2. The Evaluator uses the same 5 self-generated assertion tests. pure-MCTS uses the same search budget as the main experiments. The settings are \texttt{max\_iterations}=8, \texttt{branch\_factor}=5, \texttt{max\_depth}=8, \texttt{min\_accept\_iterations}=3, and \texttt{complete\_branch\_before\_accept}=true. The UCB exploration coefficient is $C=1.4$. pure-MCTS sets \texttt{path\_memory\_k}=0, \texttt{global\_memory\_top\_t}=0, and semantic heuristic $\alpha=\beta=0$. It also disables LLM semantic scoring and LLM probes. MCTS-style uses the same search budget as pure-MCTS. It also disables path memory, global memory, semantic UCB, and LLM probes. It enables LLM semantic scoring as a production-style reward and acceptance component. The settings are \texttt{w\_semantic}=0.3 and \texttt{semantic\_threshold}=0.45.
    \end{minipage}
    
    \scriptsize
    \setlength{\tabcolsep}{2.2pt}
    \begin{tabular}{@{}lllrrrrr@{}}
        \toprule
        Model & Dataset & Method & Hidden Solve & Accepted & Acc. Prec. & False Acc. & Oracle Pool \\
        \midrule
        \texttt{Qwen3-8B} & HumanEval & Best-of-8 & 82.93\% & 95/164 & 98.95\% & 1.05\% & 82.93\% \\
        \texttt{Qwen3-8B} & HumanEval & Reflexion & 81.10\% & 99/164 & 96.97\% & 3.03\% & 84.15\% \\
        \texttt{Qwen3-8B} & HumanEval & pure-MCTS & 84.76\% & 163/164 & 84.66\% & 15.34\% & 87.20\% \\
        \texttt{Qwen3-8B} & HumanEval & MCTS-style & 85.98\% & 90/164 & 90.00\% & 10.00\% & 86.59\% \\
        \texttt{Qwen3-8B} & MBPP & Best-of-8 & 60.42\% & 190/427 & 80.53\% & 19.47\% & 61.83\% \\
        \texttt{Qwen3-8B} & MBPP & Reflexion & 61.59\% & 234/427 & 74.36\% & 25.64\% & 64.17\% \\
        \texttt{Qwen3-8B} & MBPP & pure-MCTS & 64.87\% & 427/427 & 64.87\% & 35.13\% & 66.51\% \\
        \texttt{Qwen3-8B} & MBPP & MCTS-style & 64.87\% & 302/427 & 65.23\% & 34.77\% & 65.81\% \\
        \midrule
        \texttt{Qwen3-32B} & HumanEval & Best-of-8 & 88.41\% & 108/164 & 96.30\% & 3.70\% & 88.41\% \\
        \texttt{Qwen3-32B} & HumanEval & Reflexion & 90.24\% & 118/164 & 97.46\% & 2.54\% & 90.85\% \\
        \texttt{Qwen3-32B} & HumanEval & pure-MCTS & 89.02\% & 162/164 & 89.51\% & 10.49\% & 92.68\% \\
        \texttt{Qwen3-32B} & HumanEval & MCTS-style & 90.24\% & 125/164 & 94.40\% & 5.60\% & 92.68\% \\
        \texttt{Qwen3-32B} & MBPP & Best-of-8 & 63.00\% & 207/427 & 80.68\% & 19.32\% & 63.47\% \\
        \texttt{Qwen3-32B} & MBPP & Reflexion & 68.85\% & 289/427 & 78.55\% & 21.45\% & 70.26\% \\
        \texttt{Qwen3-32B} & MBPP & pure-MCTS & 69.79\% & 427/427 & 69.79\% & 30.21\% & 70.49\% \\
        \texttt{Qwen3-32B} & MBPP & MCTS-style & 69.56\% & 361/427 & 67.59\% & 32.41\% & 71.43\% \\
        \midrule
        \texttt{DeepSeek} & HumanEval & Best-of-8 & 95.73\% & 131/164 & 98.47\% & 1.53\% & 96.34\% \\
        \texttt{DeepSeek} & HumanEval & Reflexion & 94.51\% & 138/164 & 97.83\% & 2.17\% & 96.95\% \\
        \texttt{DeepSeek} & HumanEval & pure-MCTS & 92.07\% & 164/164 & 92.07\% & 7.93\% & 95.73\% \\
        \texttt{DeepSeek} & HumanEval & MCTS-style & 94.51\% & 162/164 & 94.44\% & 5.56\% & 95.73\% \\
        \texttt{DeepSeek} & MBPP & Best-of-8 & 73.54\% & 322/427 & 78.88\% & 21.12\% & 77.52\% \\
        \texttt{DeepSeek} & MBPP & Reflexion & 71.90\% & 356/427 & 77.81\% & 22.19\% & 74.47\% \\
        \texttt{DeepSeek} & MBPP & pure-MCTS & 70.96\% & 427/427 & 70.96\% & 29.04\% & 76.35\% \\
        \texttt{DeepSeek} & MBPP & MCTS-style & 71.19\% & 425/427 & 71.29\% & 28.71\% & 76.58\% \\
        \bottomrule
    \end{tabular}
\end{table*}

\section{Known-Feedback Baseline Results}
\label{app:known-feedback-baselines}

Table~\ref{tab:appendix-known-feedback-baselines} reports the completed known-feedback baseline results. These runs use visible tests as feedback during search and hidden tests only for final offline evaluation. Reflexion-style is a single-path Actor--Evaluator--Self-Reflection loop. MCTS-style keeps tree search but disables path memory, global memory retrieval, semantic UCB, and LLM probes.

\begin{table*}[t]
    \centering
    \caption{Known-feedback baseline completion results. Hidden Solve, Accepted, Acc. Prec., and False Acc. are percentages.}
    \label{tab:appendix-known-feedback-baselines}
    \small
    \begin{tabular}{@{}lllrrrrr@{}}
        \toprule
        Model & Dataset & Method & Hidden Solve & Accepted & Acc. Prec. & False Acc. & Avg. Iter. \\
        \midrule
        \texttt{DeepSeek-V4-Flash} & HumanEval & Reflexion-style & 96.34 & 96.34 & 100.00 & 0.00 & 1.360 \\
        \texttt{DeepSeek-V4-Flash} & HumanEval & MCTS-style & 96.34 & 96.34 & 100.00 & 0.00 & 3.183 \\
        \texttt{DeepSeek-V4-Flash} & MBPP & Reflexion-style & 94.15 & 93.21 & 100.00 & 0.00 & 1.684 \\
        \texttt{DeepSeek-V4-Flash} & MBPP & MCTS-style & 75.88 & 74.24 & 100.00 & 0.00 & 4.288 \\
        \midrule
        \texttt{Qwen3-8B} & HumanEval & Reflexion-style & 86.59 & 86.59 & 100.00 & 0.00 & 2.098 \\
        \texttt{Qwen3-8B} & HumanEval & MCTS-style & 86.59 & 86.59 & 100.00 & 0.00 & 3.445 \\
        \texttt{Qwen3-8B} & MBPP & Reflexion-style & 81.97 & 79.86 & 100.00 & 0.00 & 2.611 \\
        \texttt{Qwen3-8B} & MBPP & MCTS-style & 66.74 & 64.64 & 100.00 & 0.00 & 4.002 \\
        \midrule
        \texttt{Qwen3-32B} & HumanEval & Reflexion-style & 93.29 & 93.29 & 100.00 & 0.00 & 1.604 \\
        \texttt{Qwen3-32B} & HumanEval & MCTS-style & 93.90 & 93.90 & 100.00 & 0.00 & 3.244 \\
        \texttt{Qwen3-32B} & MBPP & Reflexion-style & 86.89 & 85.71 & 100.00 & 0.00 & 2.150 \\
        \texttt{Qwen3-32B} & MBPP & MCTS-style & 71.19 & 69.56 & 100.00 & 0.00 & 3.979 \\
        \bottomrule
    \end{tabular}
\end{table*}

\section{Logged Token Comparison}
\label{app:token-comparison}

Table~\ref{tab:appendix-token-comparison} compares the logged token use of full \remcts{} and MCTS-style on the same model--dataset--feedback settings used in the compact baseline comparison. Tokens are the sum of logged prompt and completion tokens in the experiment reports. Across all twelve comparable settings, \remcts{} uses 1.09$\times$ the total tokens of MCTS-style. The ratio is 1.14$\times$ in \notests{} and 1.04$\times$ in \known{}.

\begin{table*}[t]
    \centering
    \caption{Logged token comparison between full \remcts{} and MCTS-style. Token counts are prompt-plus-completion totals in millions (M). Ratio is \remcts{} tokens divided by MCTS-style tokens.}
    \label{tab:appendix-token-comparison}
    \small
    \begin{tabular}{@{}lllrrr@{}}
        \toprule
        Model & Dataset & Setting & \remcts{} Tok. (M) & MCTS-style Tok. (M) & Ratio \\
        \midrule
        \texttt{Qwen3-8B} & HumanEval & \notests{} & 2.52 & 2.26 & 1.12 \\
        \texttt{Qwen3-8B} & HumanEval & \known{} & 1.07 & 1.74 & 0.61 \\
        \texttt{Qwen3-8B} & MBPP & \notests{} & 4.40 & 3.96 & 1.11 \\
        \texttt{Qwen3-8B} & MBPP & \known{} & 3.93 & 3.77 & 1.04 \\
        \midrule
        \texttt{Qwen3-32B} & HumanEval & \notests{} & 2.32 & 2.23 & 1.04 \\
        \texttt{Qwen3-32B} & HumanEval & \known{} & 0.80 & 1.72 & 0.46 \\
        \texttt{Qwen3-32B} & MBPP & \notests{} & 4.16 & 3.58 & 1.16 \\
        \texttt{Qwen3-32B} & MBPP & \known{} & 4.01 & 3.52 & 1.14 \\
        \midrule
        \texttt{DeepSeek-V4-Flash} & HumanEval & \notests{} & 3.01 & 2.74 & 1.10 \\
        \texttt{DeepSeek-V4-Flash} & HumanEval & \known{} & 0.80 & 1.64 & 0.49 \\
        \texttt{DeepSeek-V4-Flash} & MBPP & \notests{} & 7.43 & 6.20 & 1.20 \\
        \texttt{DeepSeek-V4-Flash} & MBPP & \known{} & 7.66 & 5.13 & 1.49 \\
        \midrule
        \multicolumn{3}{@{}l}{\notests{} total} & 23.83 & 20.97 & 1.14 \\
        \multicolumn{3}{@{}l}{\known{} total} & 18.26 & 17.51 & 1.04 \\
        \multicolumn{3}{@{}l}{Overall total} & 42.09 & 38.48 & 1.09 \\
        \bottomrule
    \end{tabular}
\end{table*}

\section{Paired Significance Tests}
\label{app:significance}

All tests below are two-sided exact McNemar tests on paired per-task hidden outcomes. The Wilson interval is shown for the second method's solve rate. We report these tests descriptively; no claim of theoretical optimality is based on them.

\begin{table*}[t]
    \centering
    \caption{Paired hidden-test comparisons between Direct \notests{} and \remcts{} \known{}. ``Only'' columns count tasks solved by exactly one method.}
    \label{tab:appendix-main-significance}
    \footnotesize
    \setlength{\tabcolsep}{3.2pt}
    \begin{tabular}{@{}llrrrrrr@{}}
        \toprule
        Model & Dataset & Direct & ReMCTS & ReMCTS-only & Direct-only & Exact $p$ & ReMCTS 95\% CI \\
        \midrule
        \texttt{Qwen3-1.7B} & HumanEval & 99/164 & 137/164 & 40 & 2 & $4.11\times10^{-10}$ & 77.11--88.43 \\
        \texttt{Qwen3-1.7B} & MBPP & 228/427 & 271/427 & 62 & 19 & $1.77\times10^{-6}$ & 58.80--67.89 \\
        \texttt{Qwen3-8B} & HumanEval & 131/164 & 154/164 & 25 & 2 & $5.65\times10^{-6}$ & 89.14--96.65 \\
        \texttt{Qwen3-8B} & MBPP & 274/427 & 317/427 & 57 & 14 & $2.67\times10^{-7}$ & 69.89--78.16 \\
        \texttt{Qwen3-14B} & HumanEval & 136/164 & 136/164 & 15 & 15 & 1.00 & 76.43--87.92 \\
        \texttt{Qwen3-14B} & MBPP & 286/427 & 280/427 & 40 & 46 & 0.59 & 60.95--69.92 \\
        \texttt{Qwen3-32B} & HumanEval & 147/164 & 156/164 & 13 & 4 & 0.0490 & 90.67--97.51 \\
        \texttt{Qwen3-32B} & MBPP & 307/427 & 333/427 & 35 & 9 & $1.06\times10^{-4}$ & 73.82--81.66 \\
        \texttt{DeepSeek-V4-Flash} & HumanEval & 132/164 & 162/164 & 30 & 0 & $1.86\times10^{-9}$ & 95.66--99.66 \\
        \texttt{DeepSeek-V4-Flash} & MBPP & 310/427 & 346/427 & 47 & 11 & $2.03\times10^{-6}$ & 77.04--84.47 \\
        \bottomrule
    \end{tabular}
\end{table*}

\begin{table*}[t]
    \centering
    \caption{Paired comparison of \remcts{} \known{} with the same-search-budget MCTS-style baseline. MCTS-style disables path/global memory, semantic UCB, and LLM probes.}
    \label{tab:appendix-mcts-significance}
    \footnotesize
    \setlength{\tabcolsep}{3.4pt}
    \begin{tabular}{@{}llrrrrrr@{}}
        \toprule
        Model & Dataset & MCTS-style & ReMCTS & ReMCTS-only & MCTS-only & Exact $p$ & ReMCTS 95\% CI \\
        \midrule
        \texttt{Qwen3-8B} & HumanEval & 142/164 & 154/164 & 13 & 1 & 0.00183 & 89.14--96.65 \\
        \texttt{Qwen3-8B} & MBPP & 285/427 & 317/427 & 33 & 1 & $4.07\times10^{-9}$ & 69.89--78.16 \\
        \texttt{Qwen3-32B} & HumanEval & 154/164 & 156/164 & 5 & 3 & 0.727 & 90.67--97.51 \\
        \texttt{Qwen3-32B} & MBPP & 304/427 & 333/427 & 30 & 1 & $2.98\times10^{-8}$ & 73.82--81.66 \\
        \texttt{DeepSeek-V4-Flash} & HumanEval & 158/164 & 162/164 & 5 & 1 & 0.219 & 95.66--99.66 \\
        \texttt{DeepSeek-V4-Flash} & MBPP & 324/427 & 346/427 & 27 & 5 & $1.13\times10^{-4}$ & 77.04--84.47 \\
        \bottomrule
    \end{tabular}
\end{table*}

\section{Memory Ablation Results}
\label{app:memory-ablation-results}

Table~\ref{tab:appendix-memory-ablation} reports the full memory ablation results. The no-global-memory variant sets \texttt{global\_memory\_top\_t}=0. The no-path-memory variant sets \texttt{path\_memory\_k}=0. These ablations use the same final hidden-evaluation protocol as the main experiments.

\begin{table*}[t]
    \centering
    \caption{Memory ablation results. Hidden Solve, Accepted, and False Acc. are percentages.}
    \label{tab:appendix-memory-ablation}
    \small
    \scriptsize
    \setlength{\tabcolsep}{2.0pt}
    \begin{tabular}{@{}lllrrrrr@{}}
        \toprule
        Model & Dataset & Method & Hidden Solve & Accepted & False Acc. & Avg. Iter. & Avg. Tree \\
        \midrule
        \texttt{DeepSeek-V4-Flash} & HumanEval \known{} & No global & 99.39 & 99.39 & 0.00 & 3.043 & 7.165 \\
        \texttt{DeepSeek-V4-Flash} & HumanEval \known{} & No path & 99.39 & 99.39 & 0.00 & 3.030 & 7.238 \\
        \texttt{DeepSeek-V4-Flash} & HumanEval \notests{} & No global & 94.51 & 99.39 & 5.52 & 3.030 & 6.835 \\
        \texttt{DeepSeek-V4-Flash} & HumanEval \notests{} & No path & 94.51 & 100.00 & 5.49 & 3.000 & 6.695 \\
        \texttt{DeepSeek-V4-Flash} & MBPP \known{} & No global & 80.56 & 79.39 & 0.00 & 4.082 & 10.761 \\
        \texttt{DeepSeek-V4-Flash} & MBPP \known{} & No path & 80.09 & 78.45 & 0.00 & 4.082 & 11.354 \\
        \texttt{DeepSeek-V4-Flash} & MBPP \notests{} & No global & 73.07 & 99.77 & 26.76 & 3.014 & 7.862 \\
        \texttt{DeepSeek-V4-Flash} & MBPP \notests{} & No path & 72.83 & 100.00 & 27.17 & 3.000 & 7.621 \\
        \midrule
        \texttt{Qwen3-8B} & HumanEval \known{} & No global & 92.07 & 92.07 & 0.00 & 3.341 & 3.159 \\
        \texttt{Qwen3-8B} & HumanEval \known{} & No path & 95.12 & 95.12 & 0.00 & 3.213 & 3.366 \\
        \texttt{Qwen3-8B} & HumanEval \notests{} & No global & 85.37 & 87.80 & 13.19 & 3.354 & 3.262 \\
        \texttt{Qwen3-8B} & HumanEval \notests{} & No path & 86.59 & 88.41 & 12.41 & 3.518 & 3.689 \\
        \texttt{Qwen3-8B} & MBPP \known{} & No global & 69.56 & 67.68 & 0.00 & 4.124 & 3.471 \\
        \texttt{Qwen3-8B} & MBPP \known{} & No path & 73.54 & 71.43 & 0.00 & 4.389 & 4.370 \\
        \texttt{Qwen3-8B} & MBPP \notests{} & No global & 66.04 & 92.97 & 31.49 & 3.204 & 3.162 \\
        \texttt{Qwen3-8B} & MBPP \notests{} & No path & 66.51 & 94.85 & 31.60 & 3.239 & 3.290 \\
        \midrule
        \texttt{Qwen3-32B} & HumanEval \known{} & No global & 95.12 & 95.12 & 0.00 & 3.213 & 3.665 \\
        \texttt{Qwen3-32B} & HumanEval \known{} & No path & 95.12 & 95.12 & 0.00 & 3.226 & 3.811 \\
        \texttt{Qwen3-32B} & HumanEval \notests{} & No global & 90.85 & 97.56 & 8.13 & 3.098 & 3.567 \\
        \texttt{Qwen3-32B} & HumanEval \notests{} & No path & 89.63 & 98.17 & 8.70 & 3.091 & 3.677 \\
        \texttt{Qwen3-32B} & MBPP \known{} & No global & 74.71 & 72.83 & 0.00 & 4.115 & 4.052 \\
        \texttt{Qwen3-32B} & MBPP \known{} & No path & 78.69 & 77.28 & 0.00 & 4.145 & 4.934 \\
        \texttt{Qwen3-32B} & MBPP \notests{} & No global & 69.56 & 96.72 & 28.33 & 3.178 & 3.515 \\
        \texttt{Qwen3-32B} & MBPP \notests{} & No path & 70.02 & 98.59 & 28.98 & 3.098 & 3.548 \\
        \bottomrule
    \end{tabular}
\end{table*}

\section{Search-Tree Structure Statistics}
\label{app:tree}

Table~\ref{tab:appendix-tree-stats} reports search-tree statistics for all main \remcts{} experiments. For \known{}, we use the latest hidden-eval reruns. For \notests{}, we use the v2 results adopted in the main table. Effective branch is the micro-average number of child nodes over non-leaf nodes. It equals total edges divided by total non-leaf nodes.

\begin{table*}[t]
    \centering
    \caption{Search-tree structure statistics for all main \remcts{} experiments. Range gives the minimum and maximum number of tree nodes per task. Eff. Branch is the micro-average effective branch factor over non-leaf nodes.}
    \label{tab:appendix-tree-stats}
    \small
    \scriptsize
    \setlength{\tabcolsep}{2.2pt}
    \begin{tabular}{@{}llcrrrrrr@{}}
        \toprule
        Model & Dataset & Setting & Avg. Iter. & Avg. Tree & Median & Range & Avg. Depth & Eff. Branch \\
        \midrule
        \texttt{Qwen3-1.7B} & HumanEval & \notests{} & 3.098 & 5.165 & 5 & 2--17 & 1.396 & 2.858 \\
        \texttt{Qwen3-1.7B} & HumanEval & \known{} & 3.835 & 5.762 & 5 & 2--21 & 1.384 & 2.861 \\
        \texttt{Qwen3-1.7B} & MBPP & \notests{} & 3.068 & 5.433 & 5 & 2--15 & 1.412 & 2.967 \\
        \texttt{Qwen3-1.7B} & MBPP & \known{} & 4.974 & 7.651 & 7 & 2--29 & 1.609 & 2.820 \\
        \midrule
        \texttt{Qwen3-8B} & HumanEval & \notests{} & 3.500 & 3.671 & 3 & 2--12 & 1.341 & 1.904 \\
        \texttt{Qwen3-8B} & HumanEval & \known{} & 3.354 & 3.537 & 3 & 2--21 & 1.366 & 1.741 \\
        \texttt{Qwen3-8B} & MBPP & \notests{} & 3.232 & 3.389 & 3 & 2--27 & 1.241 & 1.828 \\
        \texttt{Qwen3-8B} & MBPP & \known{} & 4.361 & 4.344 & 3 & 2--22 & 1.283 & 2.112 \\
        \midrule
        \texttt{Qwen3-14B} & HumanEval & \notests{} & 3.762 & 5.793 & 5 & 2--29 & 1.360 & 2.989 \\
        \texttt{Qwen3-14B} & HumanEval & \known{} & 3.823 & 5.335 & 4 & 2--27 & 1.317 & 2.938 \\
        \texttt{Qwen3-14B} & MBPP & \notests{} & 4.162 & 6.719 & 5 & 2--37 & 1.342 & 2.942 \\
        \texttt{Qwen3-14B} & MBPP & \known{} & 4.696 & 6.260 & 5 & 2--28 & 1.279 & 2.999 \\
        \midrule
        \texttt{Qwen3-32B} & HumanEval & \notests{} & 3.079 & 3.683 & 3 & 2--27 & 1.317 & 1.956 \\
        \texttt{Qwen3-32B} & HumanEval & \known{} & 3.256 & 3.604 & 3 & 2--16 & 1.317 & 1.889 \\
        \texttt{Qwen3-32B} & MBPP & \notests{} & 3.131 & 3.628 & 3 & 2--24 & 1.283 & 1.941 \\
        \texttt{Qwen3-32B} & MBPP & \known{} & 4.204 & 5.096 & 3 & 2--22 & 1.372 & 2.141 \\
        \midrule
        \texttt{DeepSeek-V4-Flash} & HumanEval & \notests{} & 3.006 & 7.116 & 6 & 2--21 & 1.232 & 4.438 \\
        \texttt{DeepSeek-V4-Flash} & HumanEval & \known{} & 3.061 & 7.463 & 6 & 2--31 & 1.293 & 4.435 \\
        \texttt{DeepSeek-V4-Flash} & MBPP & \notests{} & 3.030 & 8.019 & 6 & 3--33 & 1.365 & 4.440 \\
        \texttt{DeepSeek-V4-Flash} & MBPP & \known{} & 4.056 & 11.593 & 8 & 3--40 & 1.564 & 4.578 \\
        \bottomrule
    \end{tabular}
\end{table*}

\section{Minimum Search Iteration Ablation}
\label{app:min-accept-ablation}

Tables~\ref{tab:appendix-min-accept-score} and~\ref{tab:appendix-min-accept} compare early experiments with the current main experiments. The early experiments set the minimum number of search iterations to 1. To make correctness metrics comparable, Score@1 for older \known{} runs is obtained by re-running hidden tests offline. These tests use the final candidate code recorded in the early run directories. Older \notests{} runs directly use the hidden solve rates in the original reports. The current \texttt{min\_accept\_iterations}=3 generally gives more stable hidden solve rates in \notests{}. In \known{}, it mainly changes search scale and leaves scores nearly unchanged.

\begin{table*}[t]
    \centering
    \caption{Effect of changing the minimum search iterations from 1 to 3 on hidden solve rate. Score@1 and Score@3 are percentages. $\Delta$Score is the percentage-point difference.}
    \label{tab:appendix-min-accept-score}
    \small
    \begin{tabular}{@{}llcrrr@{}}
        \toprule
        Model & Dataset & Setting & Score@1 & Score@3 & $\Delta$Score \\
        \midrule
        \texttt{Qwen3-1.7B} & HumanEval & \notests{} & 64.02 & 69.51 & +5.49 \\
        \texttt{Qwen3-1.7B} & HumanEval & \known{} & 82.32 & 83.54 & +1.22 \\
        \texttt{Qwen3-1.7B} & MBPP & \notests{} & 51.99 & 51.29 & -0.70 \\
        \texttt{Qwen3-1.7B} & MBPP & \known{} & 63.70 & 63.47 & -0.23 \\
        \midrule
        \texttt{Qwen3-8B} & HumanEval & \notests{} & 87.20 & 86.59 & -0.61 \\
        \texttt{Qwen3-8B} & HumanEval & \known{} & 93.29 & 93.90 & +0.61 \\
        \texttt{Qwen3-8B} & MBPP & \notests{} & 63.93 & 66.74 & +2.81 \\
        \texttt{Qwen3-8B} & MBPP & \known{} & 74.47 & 74.24 & -0.23 \\
        \midrule
        \texttt{Qwen3-14B} & HumanEval & \notests{} & 77.44 & 78.66 & +1.22 \\
        \texttt{Qwen3-14B} & HumanEval & \known{} & 82.32 & 82.93 & +0.61 \\
        \texttt{Qwen3-14B} & MBPP & \notests{} & 59.02 & 60.42 & +1.41 \\
        \texttt{Qwen3-14B} & MBPP & \known{} & 66.28 & 65.57 & -0.70 \\
        \midrule
        \texttt{Qwen3-32B} & HumanEval & \notests{} & 90.24 & 89.63 & -0.61 \\
        \texttt{Qwen3-32B} & HumanEval & \known{} & 96.34 & 95.12 & -1.22 \\
        \texttt{Qwen3-32B} & MBPP & \notests{} & 68.38 & 70.49 & +2.11 \\
        \texttt{Qwen3-32B} & MBPP & \known{} & 78.92 & 77.99 & -0.94 \\
        \midrule
        \texttt{DeepSeek-V4-Flash} & HumanEval & \notests{} & 93.29 & 95.12 & +1.83 \\
        \texttt{DeepSeek-V4-Flash} & HumanEval & \known{} & 99.39 & 98.78 & -0.61 \\
        \texttt{DeepSeek-V4-Flash} & MBPP & \notests{} & 70.26 & 72.83 & +2.58 \\
        \texttt{DeepSeek-V4-Flash} & MBPP & \known{} & 81.03 & 81.03 & +0.00 \\
        \bottomrule
    \end{tabular}
\end{table*}

\begin{table*}[t]
    \centering
    \caption{Effect of changing the minimum search iterations from 1 to 3 on search scale. Iter@1/Tree@1 come from early experiments without the current lower bound. Iter@3/Tree@3 come from the current main experiments. $\Delta$Tree is the change in average search-tree size.}
    \label{tab:appendix-min-accept}
    \small
    \begin{tabular}{@{}llcrrrrr@{}}
        \toprule
        Model & Dataset & Setting & Iter@1 & Iter@3 & Tree@1 & Tree@3 & $\Delta$Tree \\
        \midrule
        \texttt{Qwen3-1.7B} & HumanEval & \notests{} & 1.293 & 3.098 & 2.591 & 5.165 & +2.573 \\
        \texttt{Qwen3-1.7B} & HumanEval & \known{} & 2.317 & 3.835 & 3.927 & 5.762 & +1.835 \\
        \texttt{Qwen3-1.7B} & MBPP & \notests{} & 1.237 & 3.068 & 2.506 & 5.433 & +2.927 \\
        \texttt{Qwen3-1.7B} & MBPP & \known{} & 3.864 & 4.974 & 6.129 & 7.651 & +1.522 \\
        \midrule
        \texttt{Qwen3-8B} & HumanEval & \notests{} & 3.677 & 3.500 & 4.073 & 3.671 & -0.402 \\
        \texttt{Qwen3-8B} & HumanEval & \known{} & 1.585 & 3.354 & 2.677 & 3.537 & +0.860 \\
        \texttt{Qwen3-8B} & MBPP & \notests{} & 2.494 & 3.232 & 3.410 & 3.389 & -0.021 \\
        \texttt{Qwen3-8B} & MBPP & \known{} & 2.993 & 4.361 & 3.710 & 4.344 & +0.635 \\
        \midrule
        \texttt{Qwen3-14B} & HumanEval & \notests{} & 2.640 & 3.762 & 4.244 & 5.793 & +1.549 \\
        \texttt{Qwen3-14B} & HumanEval & \known{} & 2.244 & 3.823 & 3.159 & 5.335 & +2.177 \\
        \texttt{Qwen3-14B} & MBPP & \notests{} & 2.419 & 4.162 & 4.473 & 6.719 & +2.246 \\
        \texttt{Qwen3-14B} & MBPP & \known{} & 3.471 & 4.696 & 4.799 & 6.260 & +1.461 \\
        \midrule
        \texttt{Qwen3-32B} & HumanEval & \notests{} & 2.329 & 3.079 & 3.622 & 3.683 & +0.061 \\
        \texttt{Qwen3-32B} & HumanEval & \known{} & 1.354 & 3.256 & 2.470 & 3.604 & +1.134 \\
        \texttt{Qwen3-32B} & MBPP & \notests{} & 1.644 & 3.131 & 2.895 & 3.628 & +0.733 \\
        \texttt{Qwen3-32B} & MBPP & \known{} & 2.705 & 4.204 & 4.098 & 5.096 & +0.998 \\
        \midrule
        \texttt{DeepSeek-V4-Flash} & HumanEval & \notests{} & 1.104 & 3.006 & 2.463 & 7.116 & +4.652 \\
        \texttt{DeepSeek-V4-Flash} & HumanEval & \known{} & 1.116 & 3.061 & 2.530 & 7.463 & +4.933 \\
        \texttt{DeepSeek-V4-Flash} & MBPP & \notests{} & 1.014 & 3.030 & 2.136 & 8.019 & +5.883 \\
        \texttt{DeepSeek-V4-Flash} & MBPP & \known{} & 2.525 & 4.056 & 7.005 & 11.593 & +4.588 \\
        \bottomrule
    \end{tabular}
\end{table*}

\section{Qualitative Case Analysis}
\label{app:case-analysis}

The qualitative interpretation in Section~\ref{sec:ablation} is not inferred from aggregate accuracy alone. We also inspect representative search trajectories where Direct and \remcts{} differ. These cases are illustrative. They are not a full manual audit. However, they show how the same mechanisms behind the aggregate trends appear in concrete tasks.

\begin{table*}[t]
    \centering
    \caption{Representative trajectories used to interpret the aggregate trends. The examples are selected from the latest main-result runs. They use hidden tests only for offline analysis.}
    \label{tab:case-analysis}
    \small
    \begin{tabularx}{\textwidth}{@{}lX@{}}
        \toprule
        Case & Observation \\
        \midrule
        Qwen3-1.7B, HumanEval/5, \known{} & Direct fails hidden evaluation on \texttt{intersperse}. In the \remcts{} run, the search tree has 10 nodes. Three root candidates fail all visible tests. Several sibling candidates pass all 3 visible tests. The selected candidate inserts the delimiter only between adjacent elements. It passes hidden evaluation. This supports the view that weaker models often benefit from sandbox filtering and evidence-based selection. \\
        Qwen3-8B, HumanEval/26, \known{} & Direct fails hidden evaluation on \texttt{remove\_duplicates}. The \remcts{} run needs 7 iterations and reaches depth 3. Early nodes repeatedly fail visible tests. The reflection memory records two issues. The model confuses ``keep first occurrence'' with ``keep elements that occur once''. It also misses a \texttt{List} import. The final node counts element frequencies. It keeps only elements with count one. It passes all 3 visible tests and hidden evaluation. This shows how path reflections can prevent repeated repair mistakes. \\
        Qwen3-32B, HumanEval/83, \known{} & Direct fails hidden evaluation on \texttt{starts\_one\_ends}. It overcounts numbers ending in 1. Qwen3-32B Direct is already strong overall. Thus, this is a residual error rather than a broad failure mode. \remcts{} explores 10 nodes over 8 iterations. The reflection memory points to an inclusion--exclusion error and the special case $n=1$. The selected candidate passes all 7 visible tests and hidden evaluation. This supports the claim that stronger models mainly benefit by recovering a smaller set of missed implementations. \\
        Qwen3-14B, MBPP task 9, \notests{} & The task asks for the minimum positive rotation needed to obtain the same string. In \notests{}, \remcts{} accepts a candidate that returns \texttt{-1} for strings with no smaller period. Hidden evaluation expects full-length rotations, e.g., \texttt{find\_Rotations("ab")==2} and \texttt{find\_Rotations("abc")==3}. Probe evidence partially fails. Contract and static checks pass. The LLM semantic review remains optimistic. This produces a false acceptance. This case shows why proxy evidence can be insufficient on MBPP. It also shows why \notests{} improvements are less stable. \\
        \bottomrule
    \end{tabularx}
\end{table*}

\section{Evaluation Protocol Notes}
\label{app:protocol}

All tables use hidden tests as the final correctness metric. The \known{} setting allows visible test feedback during search. Therefore, its results are interpreted as test-guided mechanism validation. They are not fully equivalent to standard pass@1. The \notests{} setting uses no explicit tests. It is closer to production-style code generation without test cases. However, HumanEval and MBPP-Sanitized are still function-level benchmarks. They cannot fully cover cross-file context, dependency installation, permission control, or security risks in real software projects.

The Best-of-8, Reflexion-style, pure-MCTS, and MCTS-style baselines are run on Qwen3-8B, Qwen3-32B, and DeepSeek-V4-Flash. They test whether diverse sampling, single-path reflection, reflection-free tree search, or tree search with only LLM semantic review can explain the gains of \remcts{}. None of the four methods uses hidden tests for candidate selection. Hidden tests are used only for final offline evaluation and Oracle Pool upper-bound statistics. For pure-MCTS and MCTS-style, Oracle Pool uses non-root nodes generated in the search tree as the candidate pool.

\end{document}